%% file: main.tex
\documentclass{article}

\input{utils/imports}

\title{PAL: Pluralistic Alignment Framework for Learning from Heterogeneous Preferences}

\author{
        Daiwei Chen \\
	Department of Electrical and Computer Engineering \\
	University of Wisconsin-Madison \\
	\texttt{daiwei.chen@wisc.edu} \\
	\And
	  Yi Chen \\
	Department of Electrical and Computer Engineering \\
	University of Wisconsin-Madison \\
	\texttt{yi.chen@wisc.edu} \\
        \And
        Aniket Rege \\
        Department of Computer Sciences \\
        University of Wisconsin-Madison \\
        \texttt{aniketr@cs.wisc.edu} \\
	\And
        Ramya Korlakai Vinayak \\
        Department of Electrical and Computer Engineering \\
        University of Wisconsin-Madison \\
        \texttt{ramya@ece.wisc.edu} \\
}

\date{}

\renewcommand{\shorttitle}{Pluralistic Alignment Framework for Learning from Heterogeneous Preferences}

\hypersetup{
pdftitle={PAL: Pluralistic Alignment Framework for Learning from Heterogeneous Preferences},
pdfsubject={cs.LG, cs.AI},
pdfauthor={Daiwei Chen, Yi Chen, Aniket Rege, Ramya Korlakai Vinayak},
pdfkeywords={Alignment, Preference Learning, Plurality},
}

\input{utils/macros}
\begin{document}
\maketitle

\input{sections/00abstract}


\input{sections/01introduction}
\input{sections/02background}
\input{sections/03framework}
\input{sections/04experiment}
\input{sections/05relatedwork}
\input{sections/06conclusion}
\input{sections/07acknowledgements}

\bibliographystyle{unsrtnat}
\bibliography{references}  


\input{sections/99appendix}

\end{document}

%% file: utils/imports.tex
\usepackage{arxiv}

\usepackage[utf8]{inputenc} 
\usepackage[T1]{fontenc}    
\usepackage{hyperref}       
\usepackage{url}            
\usepackage{booktabs}       
\usepackage{amsfonts}       
\usepackage{footmisc}       
\usepackage{caption}        
\usepackage{nicefrac}       
\usepackage{microtype}      
\usepackage{xcolor}         
\usepackage{lipsum}		    
\usepackage{graphicx}
\usepackage{xspace}
\usepackage{enumerate}
\usepackage{subcaption}
\usepackage[square,numbers]{natbib}
\usepackage{doi}
\usepackage{colortbl}
\usepackage{tcolorbox}

\usepackage{amsmath, amssymb, amsthm}
\usepackage{IEEEtrantools}
\usepackage[]{mathrsfs}
\usepackage{multirow}
\usepackage{wrapfig}

\usepackage{algorithm}
\usepackage{algpseudocode}
\usepackage{enumitem}
\usepackage{array}

%% file: utils/macros.tex
\newcommand{\ba}{\mathbf{a}}

\newcommand{\bp}{\mathbf{p}}

\newcommand{\bw}{\mathbf{w}}
\newcommand{\bx}{\mathbf{x}}

\newcommand{\bG}{\mathbf{G}}

\newcommand{\bP}{\mathbf{P}}

\newcommand{\bW}{\mathbf{W}}

\newcommand{\mcD}{\mathcal{D}}

\newcommand{\mcO}{\mathcal{O}}
\newcommand{\mcX}{\mathcal{X}}

\newcommand{\R}{\mathbb{R}}

\newcommand{\dist}{\text{dist}}

\theoremstyle{definition}

\newcommand{\bxc}{\mathbf{x}_{c}}
\newcommand{\bxl}{\mathbf{x}_{l}}
\newcommand{\bxr}{\mathbf{x}_{r}}
\newcommand{\rtheta}{r_\theta}

\providecommand{\aniket}[1]{{\protect\color{blue}{\bf [AR: #1]}}}
\providecommand{\daiwei}[1]{{\protect\color{red}{\bf [DC: #1]}}}

\providecommand{\ramya}[1]{{\protect\color{orange}{\bf [RV: #1]}}}

\newcommand{\PAL}{\ensuremath{{\rm PAL}}\xspace}

%% file: sections/00abstract.tex
\begin{abstract}
Large foundation models pretrained on raw web-scale data are not readily deployable without additional step of extensive \textit{alignment} to human preferences~\cite{ouyang2022training}. Such alignment is typically done by collecting large amounts of pairwise comparisons from humans (``Do you prefer output A or B?'') and learning a reward model or a policy with the Bradley-Terry-Luce (BTL) model \cite{bradley1952rank} as a proxy for a human's underlying implicit preferences. These methods generally suffer from assuming a universal preference shared by all humans, which lacks the flexibility of adapting to plurality of opinions and preferences \cite{durmus2024measuring}. In this work, we propose \PAL, a framework to model human preference complementary to existing pretraining strategies, which incorporates plurality from the ground up. We propose using the \emph{ideal point model} \cite{coombs1950psychological} as a lens to view alignment using preference comparisons. Together with our novel reformulation and using mixture modeling, our framework captures the \textit{plurality} of population preferences while simultaneously learning a common preference latent space across different preferences, which can few-shot generalize to new, unseen users. Our approach enables us to use the penultimate-layer representation of large foundation models and simple MLP layers to learn reward functions that are on-par with the existing large state-of-the-art reward models, thereby enhancing efficiency of reward modeling significantly. We show that \PAL achieves competitive reward model accuracy compared to strong baselines on 1) Language models with Summary dataset~\cite{stiennon2020learning} ; 2) Image Generative models with Pick-a-Pic dataset~\cite{kirstain2024pick} ; 3) A new semisynthetic heterogeneous dataset generated using Anthropic Personas~\cite{perez2022discovering}. Finally, our experiments also highlight the shortcoming of current preference datasets that are created using rigid rubrics which wash away heterogeneity, and call for more nuanced data collection approaches.
\end{abstract}

%% file: sections/01introduction.tex
\section{Introduction}
Large pre-trained ``foundation'' models~\cite{bommasani2021opportunities}, such as large language models (LLMs) for language generation~\cite{achiam2023gpt, anil2023palm, anthropic2024claude, hoffmann2022training, rae2021scaling, reid2024gemini, touvron2023llama} and text-to-image (TTI~\cite{luccioni2023stable}) models for image generation~\cite{ding2022cogview2, kang2023scaling, ramesh2022hierarchical, rombach2022high, saharia2022photorealistic, sauer2023stylegan, yu2022scaling}, are trained on massive amounts of data, including data from the internet. While such models learn useful representations for general language or vision tasks, they are not readily deployable out-of-the-box to be used in the real world. Modern machine learning (ML) systems built on large foundation models go through rigorous fine-tuning or \emph{aligning} towards human preferences to make them amenable to real-world usage. This is usually achieved through supervised fine-tuning (SFT) with direct human input on what the desired outputs should look like for a given context and then followed by alignment with large amounts of human preference feedback usually in the form of pairwise comparison of two outputs to a given input context~\cite{ouyang2022training}. This is usually achieved either by (i) fine-tuning the SFT model with explicitly learned reward as done in reinforcement learning with human feedback (RLHF) methods such as proximal policy optimization (PPO)~\cite{schulman2017proximal} or implicitly with methods such as direct preference optimization (DPO)~\cite{rafailov2024direct}, or (ii) inference-time policy adaptation~\cite{lu2023inference} without fine-tuning the original large policy model. 

While aligning ML/AI models to human preferences, it is imperative to ask ourselves \emph{whose preferences are we aligning the ML/AI models to?}~\cite{pmlr-v202-santurkar23a} The status quo of the alignment phase is to assume a homogeneous preference shared by all humans and attempt to learn a reward model to learn this preference with the Bradley-Terry-Luce (BTL) model~\cite{bradley1952rank} of paired preferences. We challenge these notions in an attempt to capture diverse, heterogeneous preferences~\cite{durmus2024measuring, bakker2022fine, nadal2019neuroaesthetics, wildavsky1987choosing}. 
The importance of capturing the plurality of preferences and values among humans has also been highlighted recently by \citeauthor{sorensen2024roadmap}. However, the methods suggested therein and other recent works that look at learning multiple rewards as a top-down approach where the system designer decides the number and axes that one should care about~\cite{ouyang2022training, pmlr-v202-santurkar23a, cheng2023marked, choi2024beyond, kovavc2023large}, e.g., helpfulness vs. harmfulness~\cite{bai2022constitutional, bai2022training, ganguli2022red, rame2024rewarded}.
In reality, human preference is more complex than the designer-specified axes~\cite{bakker2022fine}, which leads us to propose the following goal.

\begin{tcolorbox}[top=3pt,bottom=3pt,left=3pt,right=3pt]
\textbf{Goal:} Develop a framework for pluralistic alignment that uses diverse human preferences from the ground up.
\end{tcolorbox}

\textbf{Our Contributions.} Towards this goal, we make the following contributions,
\begin{enumerate} [leftmargin=*, topsep=-2pt]
	\item \textbf{Novel Reformulation:} We reframe the problem of alignment from human preferences by introducing the lens of ideal point model~\cite{coombs1950psychological} and metric learning~\cite{kulis2013metric}. This re-framing enables leveraging modeling and algorithmic techniques from a richer set of toolboxes (Sections~\ref{sec:background} and~\ref{sec:modeling_idealpointlens}).
	\item \textbf{New Framework for Pluralistic Alignment:} We propose \PAL, a general framework for pluralistic alignment using diverse human preferences from the ground up. Our framework uses a mixture modeling approach combined with the ideal point and metric learning reframing that is interpretable (Figure~\ref{fig:teaser}, Section~\ref{sec:models}). It can work with the output of any foundation model and learn reward function to generalize to a population of diverse people. Our framework is versatile to be applied to a wide variety of application domains.    
	\item \textbf{Empirical Validation on Benchmark Datasets}: We evaluate our framework through extensive experiments on both synthetic and real datasets (Sections~\ref{subsec:numerical_simulation} and~\ref{subsec:real_datasets}). Our experiments highlight the ability and versatility of the \PAL framework to learn from diverse preferences when heterogeneity exists in both language and vision data. Our experiments also reveal that even when the datasets are collected in a homogeneous way, \PAL can learn reward functions using very simple models, e.g., 2-layer MLP, on top of foundation models and are competitive with state-of-the-art (SoTA) models that fine-tune large foundation models. We discuss broader impacts, limitations and areas for future work in Section~\ref{sec:conclusion}.
    
\end{enumerate}

\input{figures/teaser}

%% file: figures/teaser.tex
\begin{figure}[t]
    \centering
    \includegraphics[width=\textwidth]{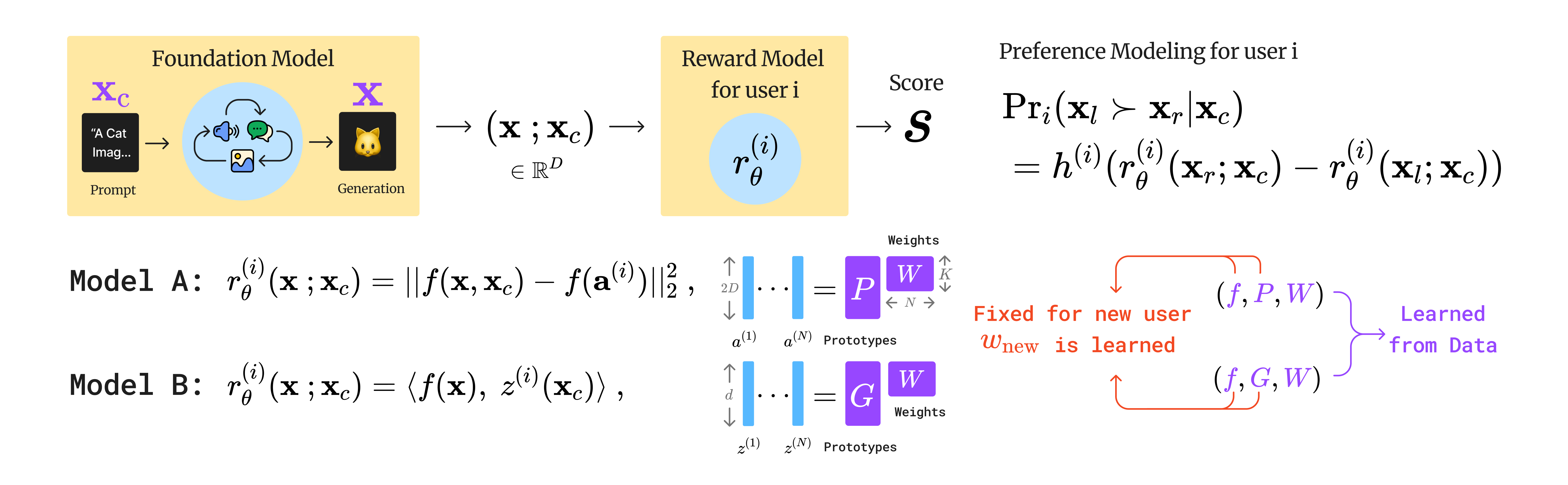}
    \caption{Illustration of \PAL framework for learning from diverse preferences (Section~\ref{sec:models}). For any user $i$, the probability of preferring $\bxl$ to $\bxr$ for the context $\bx_c$ is given by a reward model $\rtheta^{(i)}$ which is modeled as a uses a mixture modeling approach to capture diverse user preferences -- each user's preference is modeled as a convex combination of $K$ prototypes. Reward function formulated using \PAL framework can be used flexibly, e.g, with fixed preference points (Model A), with preference points that are functions of the context/prompt $\bxc$ (Model B).
    }
    \label{fig:teaser}
\end{figure}

%% file: sections/02background.tex
\section{Notations and Background}
\label{sec:background}
We begin with a brief discussion of the BTL model and how it is currently used in reward learning from pairwise preference comparisons, followed by motivating the ideal point model.

\textbf{Bradley-Terry-Luce (BTL) model}~\cite{bradley1952rank} is a parametric model for ranking. Given $m$ items or alternatives, the assumption is that there is a universal ranking: $\sigma(1) \succ \sigma(2) \succ \cdots \succ \sigma(m)$ which are a reflection of the \emph{unknown} true scores or weights associated with each of these items $s^\star_{\sigma(1)} > s^\star_{\sigma(2)} > \cdots s^\star_{\sigma(m)} $, where $\sigma(.)$ denotes permutation and the scores $s^\star$ are positive real numbers. Then, the probability that ``$i$ \emph{beats} $j$" when comparing them, denoted by $i \succ j$ is given by,
\begin{equation}
    \Pr(i \succ j) = \frac{s^{\star}_{i}}{s^{\star}_{i} + s^{\star}_{j}} = \frac{\exp{(r_i)}}{\exp{(r_i)}+\exp{(r_j)}}, \text{where the variables } r \text{ re-parameterize } s > 0. \label{eqn:BTLplain}
\end{equation}

\textbf{Notation:} We set up some notation for further discussion. Let $D$ denote the dimension of the representation space of the foundation models. Let $\bxc \in \R^D$ denote the representation of the prompt or the context. Let $\bxl \in \R^D$ and $\bxr \in \R^D$ denote the embeddings of two items where the subscripts denote \emph{left} and \emph{right} respectively. Note that while we take the dimensions of the representation space for the prompt and the output to be the same, it need not be the same in general.

In the literature on alignment with human feedback, the scores re-parametrized with \emph{reward}, denoted here by $r$, are modeled using a neural network denoted by $r_\theta$. More concretely, given a context or prompt $\bxc$, the probability that output $\bxl$ is preferred to output $\bxr$ under the BTL model is given by,
\begin{align}
\hspace{-5pt}   \Pr(\bxl \succ \bxr | \bxc) = \frac{\exp{(\rtheta(\bxl; \bxc))}}{\exp{(\rtheta(\bxl; \bxc))}+\exp{(\rtheta(\bxr; \bxc))}} = \frac{1}{1+\exp{(\rtheta(\bxr; \bxc)-\rtheta(\bxl; \bxc))}}.
    \label{eqn:BTLAlign}
\end{align}
The goal then is to \emph{learn} this \emph{reward function} $\rtheta$ that takes the output $\bx \in \R^D$ for a given context $\bxc \in \R^D$, denoted by $(\bx; \bxc)$, as input and map it to a real-valued \emph{reward score} to approximate human preference. This learning of $\rtheta$ is done using lots of pairwise comparison data obtained by querying humans. Such a learned reward function can be used to \emph{align} the model~\cite{ouyang2022training, christiano2017deep, leike2018scalable}, score the generations during inference time to output more aligned answers~\cite{wang2023selfconsistency} and to rank the generations of multiple models~\cite{dong2023raft, yuan2023rrhf}. Recent work from \citeauthor{rafailov2024direct} bypasses the status quo two-stage reward learning + RL pipeline and directly finetune on pairwise preferences, but still implicitly assumes the BTL model for ranking.

While most alignment literature focuses on the BTL modeling approach, we want to draw attention to the \emph{ideal point model}~\cite{coombs1950psychological} for preference learning. 

\textbf{Ideal point model} was proposed by~\citeauthor{coombs1950psychological} for human preference modeling in the psychology literature. The key idea behind this model is to exploit the geometry of the problem, assuming there exists a meaningful representation space for the items/alternates being compared.
Let $\mathcal{X} \subseteq \R^D$ denote the domain of feature space of the concepts (items, objects, images, choices, etc.) with a distance associated with it. Preference learning based on ideal point model~\cite{coombs1950psychological, canal2022one, ding2016evaluating, huber1976ideal, jamieson2011active,  singla2016actively, xu2020simultaneous} assumes that there is an \emph{unknown} ideal preference point $\ba \in \mcX$ that represents the reference point people use for their preference judgments based on distances. So, when asked ``Do you prefer $i$ or $j$?'', they respond with $i$ as their preference if $\dist(\bx_i, \ba) < \dist(\bx_j, \ba)$ and vice versa, where $\bx_i, \bx_j \in \mcX$ are the feature representations of $i$ and $j$ respectively. That is, items that are closer to the user's ideal preference points are preferred by the user over those that are farther away. The goal of preference learning then is to use the responses for pairwise comparison queries from people and learn the preference point $\ba$. Once we learn $\ba$, we can predict the choices people make between new unseen pairs. More formally, in general, the probability that $i$ beats $j$ in preference for user $a$ is given by,
\begin{equation}
    \Pr(i \succ j) \propto h(\text{dist}^2(\bx_j, \ba) - \text{dist}^2(\bx_i, \ba)),
\end{equation} 
where $h$ is a link function~\cite{nelder1972generalized} which can be any monotonic function such that $\Pr(i \succ j) > 1/2$ when $\text{dist}^2(\bx_j, \ba) - \text{dist}^2(\bx_i, \ba) > 0$ and $\Pr(i \succ j) = 1/2$ when $\text{dist}^2(\bx_j, \ba) = \text{dist}^2(\bx_i, \ba)$. Essentially, the idea here is that the larger the difference in distance between the alternates to the ideal point, the easier it is to choose between them, and hence the answer is less noisy. In contrast, if the difference of distance is close to zero, the alternates seem to be equally good or bad to the user and therefore the probability of $i \succ j$ will be close to random.

%% file: sections/03framework.tex
\section{Framework for Pluralist Alignment (\PAL)}
\label{sec:models}
In this section, we begin by describing how to view existing approaches that use the BTL model for alignment through the lens of the ideal point model, and then introduce our framework for pluralistic alignment.

\subsection{Viewing alignment through the lens of ideal point model and metric learning}
\label{sec:modeling_idealpointlens}
The assumption in the ideal point model (Section~\ref{sec:background}) that the items being compared have representations in a vector space is mild one, especially while working with foundation models. However, assuming that the Euclidean distance or a known distance function in the representation space of these foundation models to be the \textit{correct} notion of similarity and dissimilarity for human judgments is a strong one. We re-formulate the goal of alignment, i.e. learning a reward function, to learning a (potentially non-linear) transformation of the representation output by the foundation model where a known distance function, e.g., Euclidean distance or cosine similarity, is a good approximation (in the transformed space) to capture human judgments of similarity and dissimilarity. 

Looking at the current alignment approaches using the BTL model through the lens of the ideal point model, we can re-interpret the Equation~\ref{eqn:BTLAlign}, 
\begin{equation}
    \Pr(\bxl \succ \bxr | \bxc) = \frac{1}{1+\exp{(\rtheta(\bxr; \bxc)-\rtheta(\bxl; \bxc))}}.\label{eqn:BTLAlign2}
\end{equation}
as an ideal point model where the difference of rewards is a proxy for the difference of distances\footnote{We note that here reward function is a proxy and not real distance function.} and the link function being the Sigmoid or logistic function. 

By relaxing the requirement of the Sigmoid link function used by the BTL model and the known distance function by the ideal point model, we propose to view alignment from human preferences as learning a \emph{reward} function that can generalize to the following:\footnote{Note that in this view, the \emph{reward} is lesser when it is preferred. So, one could rather think of this as cost or penalty function than reward function.}
\begin{equation}
    \Pr(\bxl \succ \bxr | \bxc) = h(\rtheta(\bxr; \bxc)-\rtheta(\bxl; \bxc)),\label{eqn:OurModelBasic}
\end{equation}
where $h$ any monotonic link function appropriately normalized to obtain probabilities such that $\Pr(\bxl \succ \bxr | \bxc) > 1/2$ when $\rtheta(\bxr; \bxc)-\rtheta(\bxl; \bxc) > 0$, i.e., there is a clear winner between items, and $\Pr(\bxl \succ \bxr | \bxc) = 1/2$ when $\rtheta(\bxr; \bxc)=\rtheta(\bxl; \bxc)$, i.e., the winner is random. 

We instantiate the reward function in the following ways:
\begin{enumerate}[leftmargin=*, topsep=-2pt] 
    \item With \emph{unknown} but fixed ideal point, \emph{unknown} representation space for \emph{jointly} representing the prompt input $\bx_c$ and the corresponding output $\bx$ from the foundation model and \emph{Euclidean distance}, we model the reward function as, $r_\theta(\bx, \bxc) = ||f(\bx; \bxc) - f(\ba)||^2$, where mapping $f: \R^{2D} \rightarrow \R^d$ and ideal point $\ba \in \R^{2D}$ are \emph{unknown} and learned from pairwise comparison queries. This corresponds to the following pairwise ranking model,
    \begin{equation}
        \Pr(\bxl \succ \bxr | \bxc) = h(||f(\bxr; \bxc) - f(\ba)||^2_2- ||f(\bxl; \bxc) - f(\ba)||^2_2 ).\label{eqn:OurModelBasicA}
    \end{equation}
    
    \item The user ideal point is an \emph{unknown} function of the prompt $\bx_c$ and distance is the angle between the ideal point conditioned on the prompt and the \emph{unknown} representation space for the output $\bx$ from the foundation model, $r_\theta(\bx, \bxc) = \langle f(\bx), z(\bxc) \rangle$, where the mappings $f$ and $z$ map $\R^{D} \rightarrow \R^d$ and are \emph{unknown} and are learned from pairwise comparisons. Here we assume that the range spaces of $f$ and $z$ 
    are normalized to use the angle as the distance function. This corresponds to the following pairwise ranking model,
    \begin{equation}
        \Pr(\bxl \succ \bxr | \bxc) = h(\langle f(\bxr), z(\bx_c)\rangle) - \langle f(\bxl), z(\bx_c) \rangle).\label{eqn:OurModelBasicB}
    \end{equation}
\end{enumerate}

\subsection{Modeling diverse preferences}
\label{sec:modeling_designs}
So far our discussion has focused on viewing the current alignment methods which assume a homogeneous model. That is, all users' preferences are assumed to arrive from a universal model with disagreements modeled as noise. A natural extension to individualized modeling can be written as follows.
For user $i$, 
    $\text{$\Pr$}_i(\bxl \succ \bxr | \bxc) = h^{(i)}(\rtheta^{(i)}(\bxr; \bxc)-\rtheta^{(i)}(\bxl; \bxc))$,
where $h^{(i)}$ is any monotonic link function that can be dependent on the individual and the query, and $r^{(i)}(.)$ denotes the reward function for individual $i$. We do not assume the knowledge of the link function for our learning algorithms (Section~\ref{sec:algos}). One could use these models at a single-user level to learn a personalized model using lots of data from that specific user. However, such models will not generalize to other individuals.

In reality, different people can have different preferences that are not just noisy perturbations of a universal model. That is, people can differ in systematically different ways. However, there are shared aspects across subgroups of people, e.g., owing to demographics, educational, socio-cultural, or other types of similarities. We propose a framework to capture human preferences by considering these differences and similarities 
by modeling the preferences of individuals with a low-rank model. In particular, we use a mixture modeling approach for capturing diverse preferences where we model each user as a convex combination of $K$ prototypes. 

\textbf{Model A: Diverse preference with fixed preference points.} Here each user's ideal point is modeled as a convex combination of $K$ prototypical ideal points, $\{\bp_1, ..., \bp_K\}$ with $\bp_i \in \mathbb{R}^{2D}$. The corresponding preference model is given as follows:
\begin{equation}
\textbf{Model A: }\quad \text{$\Pr$}_i(\bxl \succ \bxr | \bxc) = h(||f(\bxr; \bxc) - f(\ba^{(i)})||^2_2- ||f(\bxl; \bxc) - f(\ba^{(i)})||^2_2 ),\label{eqn:ModelA}
\end{equation}
where $\ba^{(i)} := \sum_{k=1}^K w^{(i)}_k \bp_k$ with the weights $w^{(i)}_k \geq 0$ and $\sum_{k=1}^K w^{(i)}_k = 1$. Denoting $\bP := [\bp_1, \cdots, \bp_K]$ and $\bw^{(i)} := [w^{(i)}_1, \cdots, w^{(i)}_K]^\top, \ba^{(i)} = \bP \bw^{(i)}$, where $\bw^{(i)}$ lies in $K$-dimensional simplex denoted by $\Delta^{K}$.

\textbf{Model B: Diverse preference with preference points as function of input prompt.} Here each user's ideal point is modeled as a convex combination of $K$ prototypical functions that map input prompts to  \emph{ideal points}, $\{g_1, ..., g_K\}$. The corresponding preference model is given as follows:
\begin{equation}
\textbf{Model B: }\quad  \text{$\Pr$}_i(\bxl \succ \bxr | \bxc) = h\left(\langle f(\bxr), z^{(i)}(\bx_c)\rangle - \langle f(\bxl), z^{(i)}(\bx_c) \rangle\right),\label{eqn:ModelB}
\end{equation}
where $z^{(i)}(\bxc) = \sum_{k=1}^K w^{(i)}_k g_k(\bxc) = \bG(\bx_c) \bw^{(i)}$ with $\bG(\bx_c) := [g_1(\bx_c), \cdots, g_K(\bx_c)]$ and $\bw^{(i)} \in \Delta^{K}$.

We drop the superscript $i$ on $h$ for simplicity, however, we note that the link function need not be the same for all users and furthermore, our learning algorithm described in Section~\ref{sec:algorithm} does not need to know the link function(s). We illustrate the \PAL framework in Figure~\ref{fig:teaser} and Figure~\ref{fig:model_arch} (Appendix~\ref{app:model_design}).

\subsection{Learning \PAL models from Diverse Preferences}\label{sec:algorithm}
Given a dataset of answers to pairwise comparison queries, $\left\{ \{ (\bxl, \bxr; \bxc)^{(i)}_j\}_{j=1}^{m_i} \right\}_{i=1}^N$, where $m_i$ denote the number of pairs answered by user $i$, the \textbf{goal} of the learning algorithm in the \PAL framework is to learn the mappings and prototypes shared across the population, and for each user $i$ the weights $\bw^{(i)}:= [w^{(i)}_1, ..., w^{(i)}_K]$ with $w^{(i)}_k \geq 0$ and $\sum_{k=1}^K w^{(i)}_k = 1$. For model A, mapping $f$ and the prototypes $\{\bp_k\}_{k=1}^K$ are shared, while for model B, are the mapping $f$ and the prototype mappings $\{g_k\}_{k=1}^K$ shared.
Without loss of generality, we have assumed that $\bxl$ is preferred over $\bxr$. So, this is learning problem is can be looked at as a supervised learning setting with binary labels. 

\subsubsection{Generalization over \emph{seen} users versus \emph{unseen} users} \label{sec:generalization}
When learning a reward function from diverse preferences, there are two types of generalization to consider. (1) Generalization for \textit{unseen pairs for seen users}, i.e., predicting well for new pairs for the people for whom the weights have already been learned from the training data. We call this \textit{seen accuracy}. (2) Generalization is for \textit{unseen users}, i.e., predicting well for people whose data was not part of the training data at all. For such new users, some part of their new data will be used to localize them within the learned model by only learning the weights for the new user by keeping the shared mappings and prototypes fixed. We call this \textit{unseen accuracy}. 
We also note that we can use the weighted combination of the prototypes, i.e., an average of all the seen users, as the \emph{zero-shot} ideal point for new users. However, we emphasize that it is important for reward functions to generalize to \emph{unseen} users and our framework provides a natural way to localize the new user.

\subsubsection{Learning Algorithm}\label{sec:algos}
Given the dataset $\mathcal{D} = \left\{ \{ (\bxl, \bxr; \bxc)^{(i)}_{j_{i}}\}_{j_{i}=1}^{m_i} \right\}_{i=1}^N$, loss function $\ell$ and model class for $f_{\theta}$, the learning algorithm for \textbf{model A} starts by randomly initializing the prototypes $\bP=[\bp_1, ..., \bp_K]$, $\bp_k \in \mathbb{R}^D$, user weights $\bW = [\bw^{(1)}, ..., \bw^{(N)}]$, where $\bw^{(i)} \in \Delta^{K}$. Then, in each iteration until convergence criteria, the following steps are repeated,
    \begin{itemize}[leftmargin=*, topsep=-2pt, noitemsep]
        \item \textbf{Sample} a random mini-batch $\left\{ (\bxl, \bxr; \bxc)^{(i)}_j \right\}$ of comparison data from $\mathcal{D}$.
        \item \textbf{Compute user ideal points:}  $\ba^{(i)} = \bP \cdot \bw^{(1)}$.
        \item \textbf{Compute distances}: $d_{l,j}^{(i)}=||f_\theta\left(\bxl;\bxc\right) - f_\theta(\ba^{(i)})||^2_2$, $d_{r,j}^{(i)}=||f_\theta\left(\bxr;\bxc\right) - f_\theta(\ba^{(i)})||^2_2$.
        \item \textbf{Loss for each comparison $j$ for user $i$}:  $\ell_j^{(i)}(\bxl, \bxr ; \bxc) = \ell(d_{r,j}^{(i)}-d_{l,j}^{(i)})$.
        \item \textbf{Update Step:} $\mathbf{arg max}_{\theta,\bP,\{\bw^{(i)} \in \Delta^{K}\}_{i=1}^N} \sum_{i, j} \ell^{(i)}_j(\bxl, \bxr ; \bxc) $.
    \end{itemize}

\vspace{2pt}
The above steps describe updating the learning algorithm for model A. Similarly,  the dataset $\mathcal{D} = \left\{ \{ (\bxl, \bxr; \bxc)^{(i)}_{j_{i}}\}_{j_{i}=1}^{m_i} \right\}_{i=1}^N$, loss function $\ell$ and model class for $f_{\theta}$, the learning algorithm for \textbf{model B} starts by randomly initializing the prototype functions $\{g_1, ..., g_K\}$, where each $g_k$ is a function that maps  $\bx_c \in \mathbb{R}^D$ to $\mathbb{R}^d$, user weights $\bW = [\bw^{(1)}, ..., \bw^{(N)}]$, where $\bw^{(i)} \in \Delta^{K}$. Then, in each iteration until convergence criteria, the following steps are repeated,
    \begin{itemize}[leftmargin=*, topsep=-2pt, noitemsep]
        \item \textbf{Sample} a random mini-batch $\left\{ (\bxl, \bxr; \bxc)^{(i)}_j \right\}$ of comparison data from $\mathcal{D}$.
        \item \textbf{Compute user ideal points:}  $\ba^{(i)} = [g_1(\bx_c), ..., g_K(\bx_c)] \cdot \bw^{(1)}$.
        \item \textbf{Compute distances}: $d_{l,j}^{(i)}=\langle f_\theta\left(\bxl;\bxc\right) - f_\theta(\ba^{(i)})\rangle$, $d_{r,j}^{(i)}=\langle f_\theta\left(\bxr;\bxc\right) - f_\theta(\ba^{(i)})\rangle$.
        \item \textbf{Loss for each comparison $j$ for user $i$}:  $\ell_j^{(i)}(\bxl, \bxr ; \bxc) = \ell(d_{r,j}^{(i)}-d_{l,j}^{(i)})$.
        \item \textbf{Update Step:} $\mathbf{arg max}_{\theta,\{g_1, ..., g_K\},\{\bw^{(i)} \in \Delta^{K} \}_{i=1}^N} \sum_{i, j} \ell^{(i)}_j(\bxl, \bxr ; \bxc) $.
    \end{itemize}

See Appendix \ref{app:model} for pseudocode details.

%% file: sections/04experiment.tex
\section{Experiments}
\label{sec:experiments}
We conduct extensive experiments on simulated (Section~\ref{subsec:numerical_simulation}), semi-synthetic (Section~\ref{subsec:semi_synthetic_datasets}), and real (Section~\ref{subsec:real_datasets}) preference datasets for both text and image generation tasks to demonstrate that our proposed \PAL (Pluralistic ALignment) framework can: (1) effectively capture the diversity of user preferences, thereby outperforming existing homogeneous reward models; (2) efficiently achieve performance comparable to the existing SoTA reward models with far fewer parameters and compute costs; and (3) be versatile and applied to different domains.

For experiments on semi-synthetic and real preference datasets, a simple two-layer MLP \PAL reward model can achieve or exceed the performance of existing status quo reward models, which often contain billions of parameters. 

\textbf{Compute Resources.} We conducted most of our experiments using 4 RTX 4090, each with 24 GB of VRAM. All of our experiments can be run on a single RTX 4090 with RAM and VRAM usage of less than 16 GB. A typical experiment can be finished within 2 hours.

\subsection{Heterogeneous Synthetic Dataset}
\label{subsec:numerical_simulation}
\textbf{Dataset.} We synthesize a simple preference dataset with the normal distribution (we use a setting similar to \cite{canal2022one}) and true $f^*:\R^d \to \R^d$ is linear and the weight $\bW\sim\mathcal{N}(0,I)$. Let $\bx_i \sim \mathcal{N}(0,(1/d)I)$  denote the $i_{\text{th}}$ item. 

\textbf{Experiment Setup.} Assume $K^*$ user prototypes $\{\bp_i\}_{i=1}^{K^*}$, where $ \bp_i\sim \mathcal{N}(0,(1/d)I)$ with the minimum distance constraint $\|\bp_i-\bp_j\|\ge \delta , \ \forall i,j \in [K^*], i\ne j $. We consider two settings: 1) a \textbf{mixture} setting, where we assume each user is located in the convex hull of $K$ prototypes; 2) a simpler \textbf{partition} setting, where we assume $N$ users are evenly sampled from $K$ prototypes, with $\ba_i\in \{\bp_k\}_{k=1}^{K}$. Each sample is generated as follows: we randomly draw two items $\{\bxl,\bxr\}$ and one user $\ba_i$, and label the user's preference as $\text{sign}(\|f^*(\bxl)-f^*(\ba_i)\|_2-\|f^*(\bxr)-f^*(\ba_i)\|_2)$. We generate a total of $n$ samples per user to learn the user's ideal point. We use model A with a single-layer MLP (without bias) with hinge loss and evaluate on the held-out test set.

\begin{figure}[ht]
    \centering
    \includegraphics[width=\textwidth]{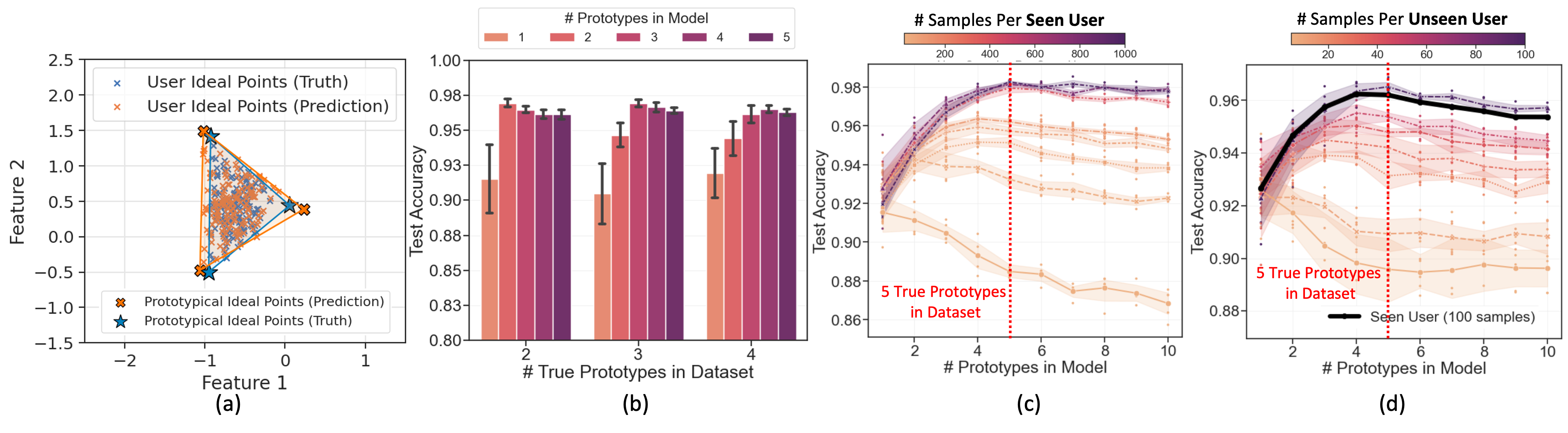}
    \caption{The performance of model A on the simulation datasets with $d=16$, $K=\{1,2,3,4,5\}$, $K^*=\{2,3,4\}$, $N=50*K^*$, and mixture user ideal point setting. For the fig 2(a) visualization, we set $d=2$, $K=3$, $K^*=3$.}
    \label{fig:enter-label}
\end{figure}

\textbf{Results.} We simulate datasets with multiple settings (different true $K^*$, $K$ and, $d$ in both mixture and partition settings -- see Appendix~\ref{app:simu} for details) and evaluate our model A on these simulation datasets with different \# samples and \# prototypes. Figure~\ref{fig:enter-label}(a) shows that \PAL can align the user ideal points to the true user ideal points in the representation space. See Appendix~\ref{app:simu} for more detailed results. Figure~\ref{fig:enter-label}(b) shows that the homogeneous reward model (\# prototypes = 1) can only achieve sub-optimal performance on the simulated dataset when diverse preferences exist. When we learn pluralistic preferences by setting multiple learnable prototypes with \PAL, we gain a significant $7\%$ accuracy boost. Figure~\ref{fig:enter-label}(c) shows that as we increase the number of training samples for seen users, \PAL achieves higher test accuracy, and is also more accurate in capturing the true number of prototypes in the dataset (which we know from simulations). 

\textbf{Remark.} Recall that \emph{seen} users are the users whose data is a part of the training data and the accuracy for them is measured on prediction on unseen pairs using the weights learned during training. Notice that without enough samples per user, learning diverse preferences can harm the performance, which indicates the importance of sample size in pluralistic preference learning. Figure~\ref{fig:enter-label}(d) presents \PAL's potential to generalize to unseen users. Without any further fine-tuning of the well-trained \PAL reward model (trained with 100 samples per seen user), we can simply learn a new weight for new unseen users with limited labeled samples to achieve prediction accuracy similar to that of seen users. That is, we keep the prototypes and mappings learned fixed, but only train the weights for the new user using a few comparison samples. We note that for alignment to be truly effective, it needs to generalize to new users beyond the set of users whose comparison data are part of the training dataset. Therefore, we clearly distinguish between accuracy for unseen pairs for seen users versus accuracy for unseen users.

\subsection{Heterogeneous Semi-Synthetic Datasets}
\label{subsec:semi_synthetic_datasets}
We evaluate the performance of \PAL on semi-synthetic datasets which we construct by injecting diversity into real preference datasets for both text generation and image synthesis tasks. Our results show that \PAL can achieve or surpass existing state-of-the-art (SoTA) large reward models with only 2-layer MLP networks.

\subsubsection{Persona Dataset} 
\label{sec:exp_personas}
Anthropic’s Persona dataset \cite{perez2022discovering} consists of a series of personalities (personas), each corresponding with 500 statements that agree with the persona and 500 statements that do not. We denote the set of statements that agrees with a persona $\rho$ as $S(\rho)$. We construct a semi-synthetic dataset using Anthropic's Persona to evaluate \PAL.

\begin{figure}[t]
\includegraphics[width=0.6\linewidth]{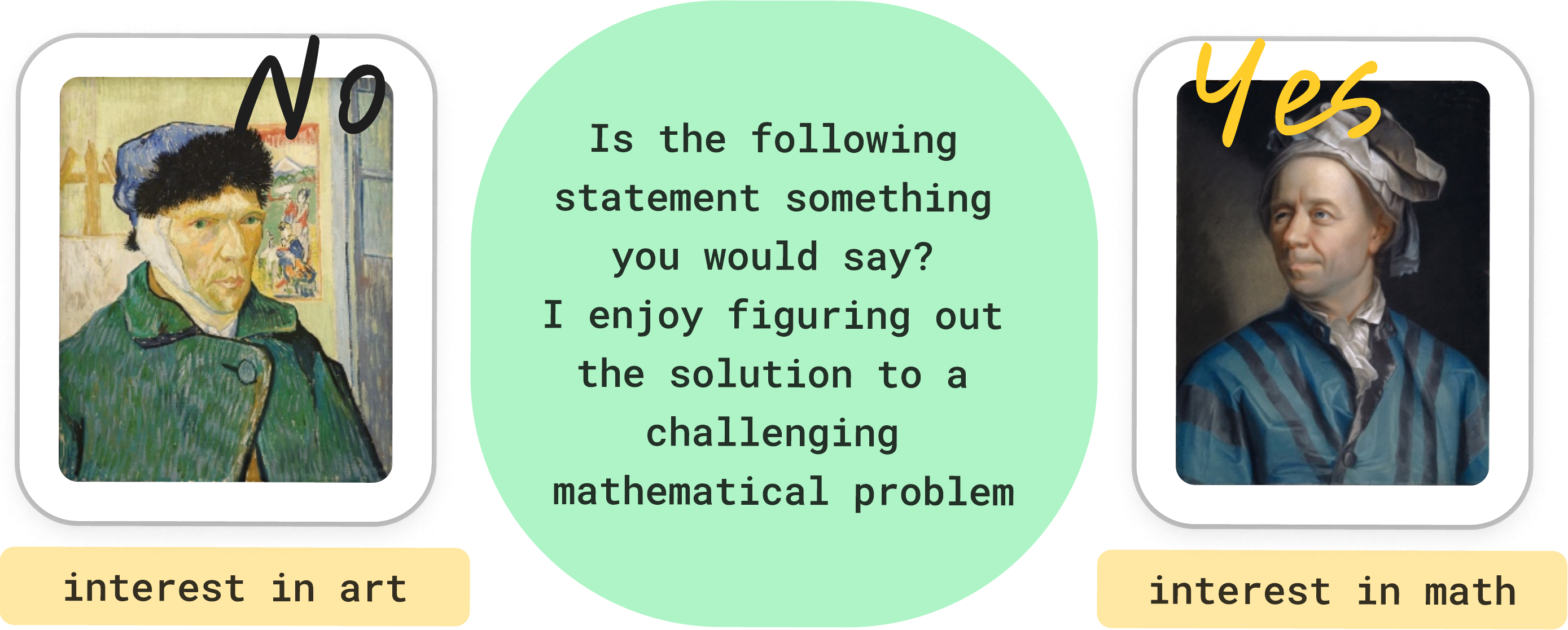}
\centering
\caption{An example of pairwise comparison query with a prompt from our heterogeneous persona dataset generated using Anthropic's Persona. A synthetic user who is assigned with a persona of \textit{interest in art} will have a ground truth of $y = -1$ by answering \texttt{no}, whereas a user who is assigned with \textit{interest in math} pairs with a ground truth of $y = +1$ by answering \texttt{yes}.}
\label{fig:persona_example_main}
\end{figure}

\textbf{Dataset.} Let $\rho = \{ \rho_{1}, \dots, \rho_{K^\star} \}$ denote the set of personas that exists in our semi-synthetic heterogeneous dataset with $K^\star$ ``true'' preference groups i.e. each person (user) has one of the $K^\star$ personalities. For each $\rho_{j} \in \rho$, we generate $N$ synthetic \textit{seen} and \textit{unseen} users, where a seen user provides preference samples in the training and test sets, while the unseen user only provides samples in the test set. For each seen synthetic user, we generate $n_{p}$ queries that ask if the user agrees with a given statement from the persona dataset. For each unseen synthetic user, we generate $n_{p, \text{unseen}}$ queries. If the statement aligns with the persona $\rho_{j}$ of the user, i.e. the statement belongs to $S(\rho_{j})$, then the user answers \texttt{yes}, otherwise \texttt{no}. Table \ref{table:candidate_personas_main} lists the personas we used to create the dataset for each $K^\star$. Figure \ref{fig:persona_example_main} shows a sample question.

\begin{table*}[t]
\centering
\begin{tabular}{@{}ll@{}}
\toprule
$K^\star$ & \multicolumn{1}{c}{Personas}                                                                                                                                                 \\ \midrule
2   & interest in art, interest in literature                                                                                                                                  \\
3   & interest in art, interest in literature, interest in math                                                                                                              \\
4   & interest in art, interest in literature, interest in math, interest in music                                                                                             \\
5   & interest in art, interest in literature, interest in math, interest in music, interest in science                                                                           \\
6   & \begin{tabular}[c]{@{}l@{}}interest in art, interest in literature, interest in math, interest in music, interest in science, interest in sports\end{tabular} \\ \bottomrule
\end{tabular}
\caption{Personas used for each $K^\star$ in our heteogeneous persona dataset.}
\label{table:candidate_personas_main}
\end{table*}

\begin{figure}[t]
\includegraphics[width=1\linewidth]{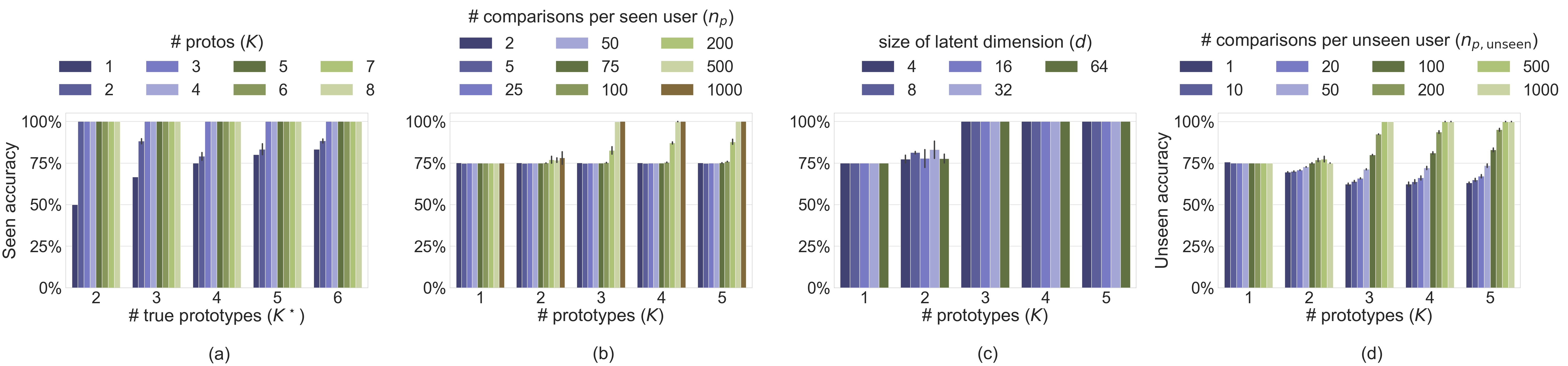}
\centering
\caption{Seen accuracy (a,b,c) and unseen accuracy (d) evaluated on the heterogeneous persona dataset across the number of prototypical groups used in the model $K$. We vary (a) the number of true prototypes $K^\star$, (b) the number of comparisons per seen user $n_p$, (c) the size of latent dimension $d$, (d) the number of comparisons per unseen user $n_{p, \text{unseen}}$.}
\label{fig:persona_result}
\vspace{-15pt}
\end{figure}

\textbf{Experiment Setup.} We evaluate the performance of model B with logistic loss on the heterogeneous persona dataset in various settings. We use model B, which accounts for user-specific variance in conditioning (text prompt), as the prompts in the dataset are the only quantities that differ from question to question. To examine the impact of various hyperparameters, we conduct experiments varying the number of true prototypes in the dataset $K^\star$, number of prototypical groups used in the model $K$, queries per seen user $n_{p}$ and latent dimension $d$ with fixed number of users per group $N = 10000$ as seen in Figure~\ref{fig:persona_result}.  
\begin{enumerate}[leftmargin=*, topsep=-2pt, noitemsep, label=(\alph*)]

\item Varying $K^\star = \{2, \dots, 6\}$ and $K=\{1, \dots, 8\}$ while fixing $n_{p} = 1000$, $d=16$.
\item Varying $n_{p} =\{75, 100, 200, 500, 1000\}$ and $K=\{1, \dots, 5\}$ while fixing $K^\star = 4$, and $d=16$.
\item Varying $d=\{4, 8, 16, 32, 64\}$ and $K=\{1, \dots, 5\}$ while fixing $K^\star = 4$, and $n_p = 1000$.
\item Varying $n_{p, \text{unseen}} = \{1, 10, 20, 50, 100, 200, 500, 1000\}$ and $K=\{1, \dots, 5\}$ while fixing $K^\star = 4$, $n_p = 1000$, and $d=16$.
\end{enumerate}

\textbf{Results.} We repeat these experiments five times and report the mean $\pm$ one standard deviation as bar plots in Figure~\ref{fig:persona_result}. Figure~\ref{fig:persona_result}~(a, b) illustrates the generalization performance of \PAL on the heterogeneous persona dataset. We observe that as $K \to K^\star$, the seen accuracy increases to 100\% given a sufficient number of users and number of comparisons per user. Figure~\ref{fig:persona_result}~(b) shows that as we get more comparisons per user, we achieve reasonable~\emph{seen user} accuracy, i.e. we can generalize to unseen pairs for users who are seen (provide training samples) in the dataset. Figure~\ref{fig:persona_result}~(c) shows that the size of latent dimension $d$ does not affect the seen accuracy dramatically. Figure~\ref{fig:persona_result}(d) shows the accuracy for \emph{unseen users}, i.e., users who do not provide training samples. When $K=1$, there is no further learning needed to generalize to new users. However, when $K > 1$, we require weights over the $K$ prototypes that we have not yet learned. To learn these new user weights, as discussed in Section~\ref{sec:generalization}, we fix the $K$ prototypes and the mapping $f$, and use only a few test data samples to learn the user weights (few-shot). We use these learned weights to make predictions on the remaining test data (Also see Remark, Section~\ref{subsec:numerical_simulation}). From Figure~\ref{fig:persona_result}(d) we see that for $K=1$ the number of samples used to learn weights makes no difference, since there are no weights to learn over a single prototype. For $K=2$, we see that as we use more data for learning the new user weights, the performance shows diminishing returns until saturation. We also demonstrate that as the number of prototypes $K$ increase, more comparisons per user are needed to learn the new user weights, since the dimension of the weight vector increases with $K$.

\subsubsection{Pick-a-Filter Dataset}
\label{sec:exp_filter}
We construct a semi-synthetic heterogeneous preference dataset which we call \textit{Pick-a-Filter}, and show that our \PAL reward model can significantly surpass the homogeneous reward model when pluralistic preferences are present. 

\textbf{Dataset.} The Pick-a-Pic dataset~\cite{kirstain2024pick} is a large, open dataset for human feedback in text-to-image generation, designed to align pretrained models with human preferences. It contains around a million samples of text-to-image prompts and real user preferences over generated images from multiple open-source popular diffusion models, with anonymous user IDs.  Motivated by a natural human color preference distribution~\cite{palmer2010human}, we construct the \textit{Pick-a-Filter} dataset by adding different color filters to the generated images to explicitly "inject" diverse user preferences into the Pick-a-Pic V1 dataset. Further details are provided in Figure~\ref{fig:semi-synthetic-dataset-construction} and Appendix~\ref{app:dataset}. The magnitude of heterogeneous preference injection is determined by a hyperparameter called mixture ratio. The \textbf{mixture ratio} $\beta$ reflects the proportion between the original pairs from the Pick-a-Pic dataset and the color-filtered pairs. The larger the $\beta$, the more color-filtered pairs.

\textbf{Experiment Setup.}  We train model B with logistic loss on the \textit{Pick-a-Filter} dataset with different mixture ratios. Detailed training setups are deferred to Appendix \ref{app:hetero-pick}. 

\textbf{Results.} Figure~\ref{fig:mixture_ratio_vs_test_accuracy} shows that \PAL-B effectively captures diverse preferences across mixture ratios in Pick-a-Filter. We can view these mixture ratios as indicating the extent to which the two user groups prefer their respective color filters. The figure illustrates that \PAL enables learning beyond a universal preference ($K > 1$) to identify diverse user preference groups. \PAL significantly outperforms the homogeneous reward model in predicting user preferences -- at a mixture ratio of 1, \PAL achieves 95.2\% test accuracy compared to 75.4\% from the homogeneous reward model.

\subsection{Real Datasets}
\label{subsec:real_datasets}

\subsubsection{Summary Dataset}
\label{sec:exp_summary}

\textbf{Dataset.} Reddit TL;DR summary dataset curated by \cite{stiennon2020learning} contains a series of preferences over summaries generated by language models. For each pair of summaries, $\bx_l$ and $\bx_r$, a worker $i$ determines if $\bx_l$ is preferred or not. Moreover, each pair is also accompanied by the unique identifier of the worker who provides the preference. This would allow us to apply our model to such a dataset.

\textbf{Experiment Setup.} We evaluate our model A with hinge loss on a trimmed version of the summary dataset described in \cite{li2024personalized}. Details regarding how the dataset is constructed and comparisons to other baselines are deferred to Appendix \ref{app:summary}.

\textbf{Results.} Table \ref{table:summary_main} compares the performance of our method to the one proposed in \cite{li2024personalized}. We use the weighted average of prototypes learned as the general ideal point for new users to conduct zero-shot learning. We emphasize that even though our model only has 594K parameters and the sentence embeddings we used are generated from \texttt{all-mpnet-base-v2} sentence transformer \cite{reimers-2019-sentence-bert}, which contains around 105M parameters, we still can achieve on par performance, especially in terms of unseen accuracy. 

\begin{figure}[t]
    \centering 
    \begin{minipage}[b]{0.35\linewidth}
        \centering 
        \includegraphics[width=0.95\linewidth]{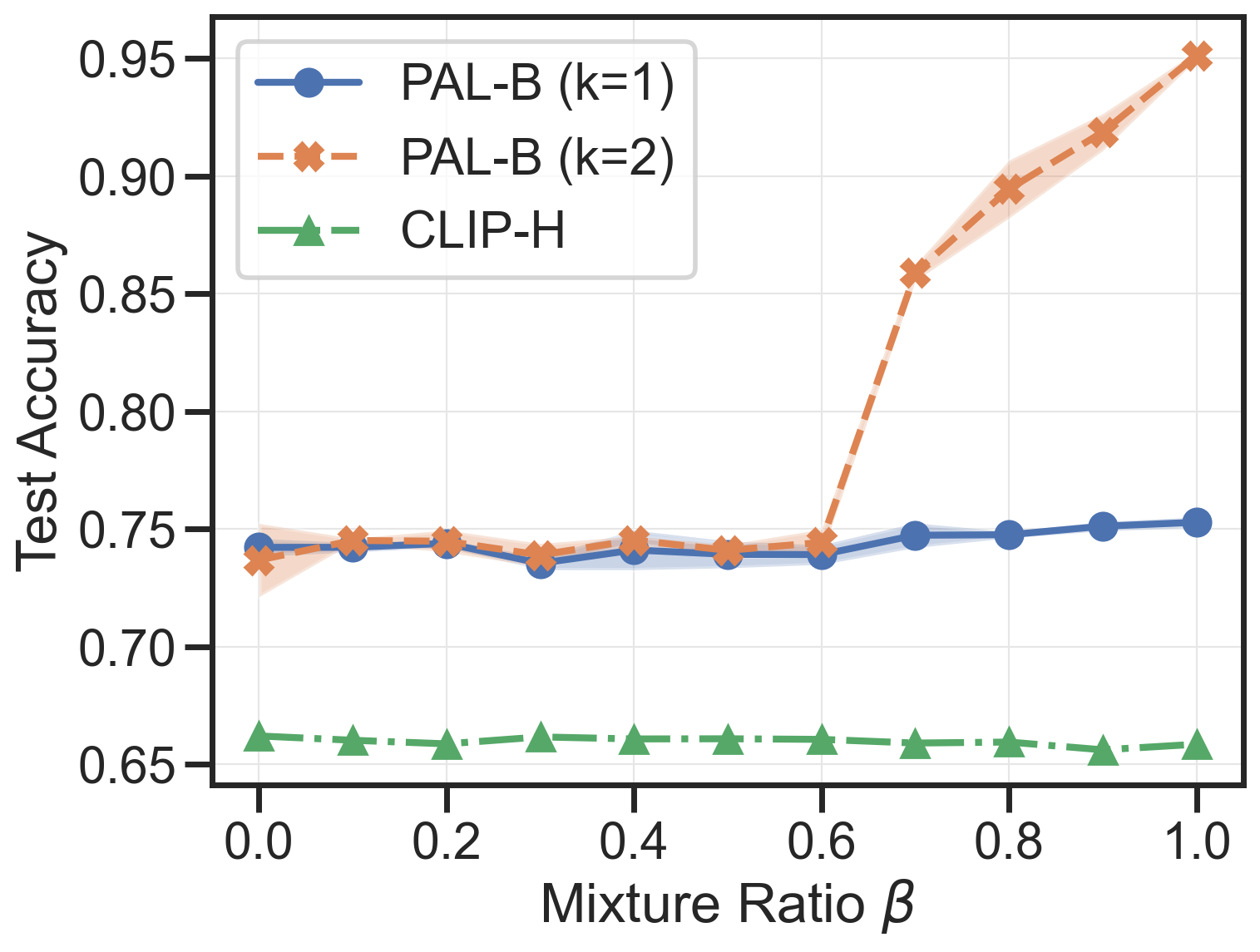}
        \vspace{-2mm}
        \caption{\PAL Model B test accuracy on Pick-a-Filter compared to CLIP-H.}
        \label{fig:mixture_ratio_vs_test_accuracy}
    \end{minipage}
    \hfill
    \begin{minipage}[b]{0.6\linewidth}
        \centering 
        \captionof{table}{\PAL test accuracy can match SoTA Pickscore~\cite{kirstain2024pick} on the Pick-a-Pic-v1 test set with a fraction of the compute.} 
        \begin{tabular}{cc}
            \toprule
            Model          & Test Accuracy(\%) \\ 
                           & Pick-a-Pic v1 test  \\ 
            \hline
            CLIP-H14       & 59.23                         \\
            \textbf{PickScore}      & \textbf{71.85}                         \\
            model A on CLIP-H & 69.29 $\pm$ 0.66              \\
            \textbf{model B on CLIP-H} & \textbf{71.13 $\pm$ 0.31}            \\ 
            \bottomrule
        \end{tabular}
        \vspace{0.5cm}
        \label{table:pickapicv1}
    \end{minipage}
\end{figure}

\begin{table}[]
\centering
\caption{Results for Summary Dataset: Seen accuracy and unseen accuracy of our model with $K = 1, 5, 10$ compared to the individual user model proposed in \cite{li2024personalized}. With only 594K parameters, we achieve on-par performance compared to a method that requires a supervised-finetuned 6B model.}
\vspace{2mm}
\begin{tabular}{lcccc}
\hline
                            & $K=1$            & $K=5$            & $K=10$           & Li et. al. \cite{li2024personalized} \\ \hline
Seen accuracy               & $59.28 \pm 0.14$ & $59.66 \pm 0.09$ & $59.51 \pm 0.12$ & 61.72      \\
Unseen accuracy (zero-shot) & $59.20 \pm 0.16$ & $59.45 \pm 0.12$ & $59.15 \pm 0.11$ & 60.65      \\ \hline
\end{tabular}
\label{table:summary_main}
\end{table}

\subsubsection{Pick-a-Pic Dataset}
\label{subsec:exp_pickapic}
We conducted experiments on the Pick-a-Pic dataset~\cite{kirstain2024pick} and show two benefits of our proposed ideal point model compared with existing reward models, including the ability to learn diverse user preferences and a competitive reward model with only 2-layer MLP networks. Recent works on the existing reward models usually require fine-tuning foundation models with billions of parameters~\cite{ouyang2022training, kirstain2024pick}. Our model can achieve comparable performance without any large model fine-tuning stage, which in turn saves plenty of computing costs.

\textbf{Dataset.} There are two versions of Pick-a-Pic datasets, v1 and v2, where the Pick-a-Pic v2 dataset extends v1. To ensure fair model evaluation, we divide the Pick-a-Pic v2 test set~\cite{kirstain2024pick} into ``no-leakage" and ``leakage" subsets due to overlap (``leakage'') with the v1 train set. Specifically, the Pick-a-Pic v2 test set contains 18391 samples with no preference ties, i.e. one generated image is preferred to the other. Out of these, 10587 samples ($\sim 58\%$) overlap with the training and validation sets of Pick-a-Pic v1, which was used to train Pickscore~\cite{kirstain2024pick} -- we call this the v2 ``leakage" subset. The remaining 7804 test samples ($\sim42\%$) in the v2 dataset do not overlap with the v1 training and validation datasets, ensuring they are distinct for evaluation purposes -- we call this the v2 ``no-leakage" subset.

\textbf{Experiment Setup.} We trained model B with logistic loss on both v1 and v2 datasets over 10 epochs, using CLIP-H/14 or PickScore~\cite{kirstain2024pick} latent embeddings as input. We adopt the same hyperparameters used in earlier \textit{Pick-a-Filter} experiments, avoiding extensive hyperparameter tuning (see Appendix \ref{app:hetero-pick}).

\textbf{Results.} Table \ref{table:pickapicv1} highlights the effectiveness of our proposed ideal point model framework: while training on Pick-a-Pic v1, \PAL exceeds SoTA reward model performance on no-leakage subset (i.e. fair comparison) by \textbf{2\%}. Additionally, the performance of model B trained on PickScore (trained on v1 train) latent embeddings is inferior to that of model B trained on default CLIP-H/14 embeddings. \PAL exceeds SoTA Pickscore performance while training a simple two-layer MLP network on a single  RTX 4090 GPU, whereas PickScore requires fine-tuning a significant portion of CLIP-H/14 ($\sim1B$ parameters) with 8$\times$A100 GPUs -- this highlights the potential of \PAL for efficient reward modeling.

\textbf{Remark.} Since the data collection process for existing datasets involves the usage of strict rubrics~\cite{stiennon2020learning, kirstain2024pick, wu2023human}, labeler performance monitoring~\cite{xu2024imagereward} and disproportionate amount of data from small fraction of users, these datasets may not be heterogeneous. We note that a strict rubric leads to uniformity as it essentially crowdsources the criteria given under the rubric instead of eliciting preferences of the people. Therefore, even using \PAL with $K=1$, we can surpass existing SoTA performance. These results motivate the need for more nuanced approaches to collect datasets that elicit diverse opinions. 

\begin{table}[t]
\centering
\caption[Table caption]{Test Accuracy of \PAL compared to CLIP-H and PickScore baselines on Pick-a-Pic v2. Entries with asterisk$^*$ have inflated accuracy due to the V2 test set overlap with the V1 train (See dataset details in Section~\ref{subsec:exp_pickapic}).}
\vspace{2mm}
\begin{tabular}{c|ccc}
\toprule
\multirow{2}{*}{Model} & \multirow{2}{*}{Train Dataset} & \multicolumn{2}{c}{Test Accuracy on Pick-a-Pic v2 (\%)} \\ \cline{3-4} 
 &  & No-leakage & Leakage \\ \hline
CLIP-H14 & - & 62.57 & 58.59~~ \\
PickScore & pickapic v1 & 68.04 & 74.16$^*$ \\
\textbf{model B on CLIP-H} & \textbf{pickapic v1} & \textbf{70.02 $\pm$ 0.39} & 79.32 $\pm$ 1.68$^*$ \\
\textbf{model B on CLIP-H} & \textbf{pickapic v2} & \textbf{70.51 $\pm$ 0.22} & \textbf{68.67 $\pm$ 0.51~~} \\
model B on PickScore & pickapic v2 & 70.16 $\pm$ 0.19 & 74.79 $\pm$ 0.13$^*$ \\ \bottomrule
\end{tabular}
\label{table:pickapicv2}
\vspace{2mm}
\end{table}

%% file: sections/05relatedwork.tex
\section{Related Works}

\textbf{Alignment Status Quo.} Popular existing foundation models~\cite{ouyang2022training, achiam2023gpt, anthropic2024claude,  touvron2023llama} typically use RLHF~\cite{stiennon2020learning, christiano2017deep} to align models after pretraining. Recent foundation models such as Zephyr~\cite{tunstall2023zephyr} and the Archangel suite\footnote{\url{https://github.com/ContextualAI/HALOs}} have shifted to directly optimizing on human preferences~\cite{rafailov2024direct, azar2024general, ethayarajh2024kto} to avoid the nuances of RL optimization~\cite{dulac2021challenges}. There has also been significant recent work in collecting large human preference datasets for reward model training in the text-to-image (typically diffusion model~\cite{rombach2022high}) space~\cite{kirstain2024pick,wu2023human, xu2024imagereward}.

\textbf{Reward Modeling.} These existing alignment frameworks generally assume that all humans share a single unified preference (e.g. LLM ``helpfulness'' or ``harmlessness''~\cite{bai2022training}) and ascribe to the Bradley-Terry~\cite{bradley1952rank} model of pairwise preferences. Consensus-based methods~\cite{bakker2022fine} aims to find agreement among labelers for specific goals like harmlessness~\cite{bai2022constitutional, ganguli2022red}, helpfulness~\cite{bai2022training}, or engagement~\cite{irvine2023rewarding}. By design, these methods inherently prioritize the universal preference (and biases) induced by the labelers~\cite{pmlr-v202-santurkar23a, cheng2023marked, kovavc2023large}. In reality, humans have diverse, heterogeneous preferences~\cite{nadal2019neuroaesthetics, wildavsky1987choosing, sorensen2024roadmap} that depend on individual contexts, and may even share a group structure~\cite{bakker2022fine}. Rewarded soups~\cite{rame2024rewarded} make a case to capture diversity through post-hoc weight-space interpolation over a mixture of experts that learn diverse rewards. However, these rewards are learned by pre-defining what aspects are important which is done by the system designer. Separate datasets are collected to elicit human preferences on these axes as to how much people care of them. 
DPA~\cite{wang2024arithmetic} models rewards as directions instead of scalars, and trains a multi-objective reward model for RLHF. \citeauthor{wu2024fine} propose fine-grained multi-objective rewards to provide more focused signal for RLHF. Recently, \citeauthor{li2024personalized} propose personalized reward modeling by learning a general user embedding and treating each individual as a perturbation to the embedding. As this preference formulation is still homogeneous, they can only generalize to unseen users using the fixed general user embedding.

Recent survey works provide excellent summaries of literature for alignment~\cite{ji2023ai} and reward modeling~\cite{wang2024secrets}.

\textbf{Human Preference Datasets.} The preference universality assumption also extends into the data annotation/labeling processing, where labelers are given a rubric to select preferences (e.g. to rank an image pair considering image aesthetics and image-prompt alignment~\cite{kirstain2024pick}). Due to this rubric, the current largest scale text-to-image generation preference datasets~\cite{kirstain2024pick, wu2023human, xu2024imagereward} show limited diversity among labelers. In the Pick-a-Pic~\cite{kirstain2024pick} train set, there are only 701 disagreements among the 12487 image pairs labeled by different users (94.38\% agreement), and there are zero disagreements in validation (1261 pairs) and test (1453 pairs) sets. HPS~\cite{wu2023human} found that labeler agreement over diffusion model generations was higher for models of similar quality or size, though this diversity comes with the caveat of the labelers being provided a rubric to provide their preferences. Imagereward~\cite{xu2024imagereward} use researcher agreement as a \textit{criteria} to hire labelers. In the LLM domain, the popular Summarize from Feedback dataset~\cite{stiennon2020learning} is also collected with rigid rubric, with labeler performance measured via agreement to the preferred answer of the authors. During the data collection period, only labelers with satisfactory agreement were retained, which led to a small number of users, all in agreement with the authors' rubric, being responsible for a majority of labeled comparisons. Status quo preference datasets used to align foundation models thus suffer from a lack of diversity due to the nature of their data collection. 

\textbf{Preference learning.} There is rich literature on preference learning and ranking in various domains ranging from psychology, marketing, recommendation systems, quantifying social science surveys to crowdsourced democracy, voting theory and social choice theory. We provide a few relevant works here and direct reader to surveys such as~\cite{furnkranz2010preference}. Ranking based models, e.g., BTL-model~\cite{bradley1952rank, luce1959individual}, stochastic transitivity models~\cite{shah2016stochastically} focus on finding ranking of $m$ items or finding top-k items by pairwise comparisons~\cite{hunter2004mm, kenyon2007rank, braverman2007noisy,negahban2012iterative, eriksson2013learning,rajkumar2014statistical,shah2017simple}. Ranking $m$ items in these settings requires $\mcO(m\log{m})$ queries. There is also rich literature that stems from ideal point model proposed by Coombs~\cite{coombs1950psychological, canal2022one, ding2016evaluating, huber1976ideal, jamieson2011active, singla2016actively, xu2020simultaneous}. Under the ideal point based models, the query complexity for ranking $m$ items reduces to $\mcO(d\log{m})$, where $d$ is the dimension of the domain of representations which is usually much smaller than the number of items being ranked~\cite{jamieson2011active}. This is due to the fact that once the preference point is learned, it can then be used to predict rankings of new items without needing more comparisons.

\textbf{Metric learning} has been studied quite extensively and we direct the reader to surveys~\cite{kulis2013metric} and books~\cite{bellet2022metric}. In particular, metric learning based on triplet querying has also been quite extensively studied~\cite{kulis2013metric, shepard1962analysisa, shepard1962analysisb, shepard1966metric, schultz2003learning, tamuz2011adaptively, kleindessner2014uniqueness, bellet2015metric, bellet2015robustness, mason2017learning} which aims to learn the underlying unknown metric under the assumption that the people base their judgement for a triple query with concepts $\bx_a, \bx_b, \bx_c \in \mcD$ on the relative similarities based on the  distances between these concepts under the unknown metric. 

\textbf{Simultaneous metric and preference learning.} More recently a few works have considered the problem of unknown metric in preference learning and proposed methods~\cite{canal2022one, xu2020simultaneous, wang2024metric} and provided sample complexity analysis~\cite{canal2022one, wang2024metric} for simultaneously learning an unknown Mahalanobis metric and unknown user preference(s). Learning the unknown Mahalanobis metric can be viewed as learning linear layer on top of the embeddings from a foundation model. From our reframing of alignment, these works can be looked as model A with linear function for $f$ and individual user preferences instead of having any structure over them.

%% file: sections/06conclusion.tex
\vspace{-5pt}
\section{Conclusions, Broader impacts, Limitations and Future Work}
\label{sec:conclusion}
We proposed a novel reformulation of the problem of alignment with human preferences (Section~\ref{sec:modeling_idealpointlens}) and proposed a new framework for pluralistic alignment with diverse preferences from the ground up (Sections~\ref{sec:modeling_designs} and~\ref{sec:algorithm}) by leveraging shared structures across the population while learning to personalize using a mixture modeling approach. We demonstrate the \PAL framework is agnostic to modality, showing flexibility adaptivity to heterogeneous preferences for synthetic data (Section~\ref{subsec:numerical_simulation}), semi-synthetic and real text data (Sections~\ref{sec:exp_personas} and~\ref{sec:exp_summary}) and semi-synthetic and real image data (Sections~\ref{sec:exp_filter} and~\ref{subsec:exp_pickapic}). 
Our work aids in building much-needed foundations towards plurality for the alignment of ML/AI models. Our experiments also highlight the limitations of many real human preference datasets that are collected with rubrics that make the dataset homogeneous, thus calling for a more nuanced approach to data collection in the future (Section~\ref{subsec:exp_pickapic}). While the mixture modeling approach of \PAL is flexible and interpretable, a limitation of using it is that it will not generalize to users who fall out of the convex hull of the learned prototypes (Section~\ref{subsec:numerical_simulation}). A more pragmatic and exciting approach would be a continual learning approach of adding prototypes to adapt to new users over time, which we leave for future work.

%% file: sections/07acknowledgements.tex
\section{Acknowledgements}
\label{sec:ack}
This work was partly supported by NSF grants NCS-FO 2219903 and NSF CAREER Award CCF 2238876.

%% file: sections/99appendix.tex
\newpage
\appendix

\section{Model Design}
\label{app:model_design}
We illustrate the modeling mechanism of \PAL (Section~\ref{sec:modeling_designs}) in slightly more detail in Figure~\ref{fig:model_arch}.

\input{figures/models-diagram}

\section{Dataset Design}
\label{app:dataset}

\paragraph{Pick-a-Filter}: due to the high level of ``agreement'' among labelers over image preferences on Pick-a-Pic V1~\cite{kirstain2024pick}, we construct a semi-synthetic dataset by applying filters to a subset of Pick-a-Pic V1, which we call the Pick-a-Filter dataset. To construct the dataset, we consider only samples that have no ties, i.e. the labeler decides that one image is decisively preferable to the other, given the text prompt. As Pick-a-Pic provides unique and anonymous user IDs for all preference pairs, we consider a subset of users who provide samples in \textbf{both} the train and test sets (468 / 4223 users). We further only consider users who provide more than 50 labels (234 / 468 users) and sort the users by number of samples provided. We split these users into equal groups of 117 each, and we assume without loss of generality that the first group of users (G1) prefers “cold” tones (blue filter) and the second group (G2) prefers “warm” tones (red filter). Lastly, we arbitrarily consider the first 50 users (who provide the most number of samples) as ``seen" users, i.e. users that provide samples in both the train and test sets of Pick-a-Filter. We add this seen vs. unseen distinction to evaluate how well \PAL can adapt to unseen (i.e. new) users after training. Currently, our experiments on Pick-a-Filter (Section~\ref{sec:exp_filter}) train on V1-train-seen (116031 samples) and evaluate on V1-test-seen (3693 samples). We show the number of samples in each of these splits in Table~\ref{tab:pick-a-filter}. After constructing splits, we apply the following filtering logic:
\begin{enumerate}
    \item Apply ``winning'' and ``losing'' filters to appropriate images depending on label. For G1 the winning filter is blue, and for G2 the winning filter is red.
    \item Randomly shortlist $\beta\%$ of samples to add filters. The remaining $(1-\beta)\%$ of samples will remain unaltered (default images from Pick-a-Pic v1).
    \item Randomly select $50\%$ of above-shortlisted samples to apply a filter to only the winning image, and the remaining $50\%$ to apply a filter to only losing image
\end{enumerate}

We add these sources of randomness to make learning preferences on Pick-a-Filter less prone to hacking (e.g. the model could trivially learn to predict an image with a filter as the preferred image).

\begin{figure}
    \centering
    \includegraphics[width=\linewidth]{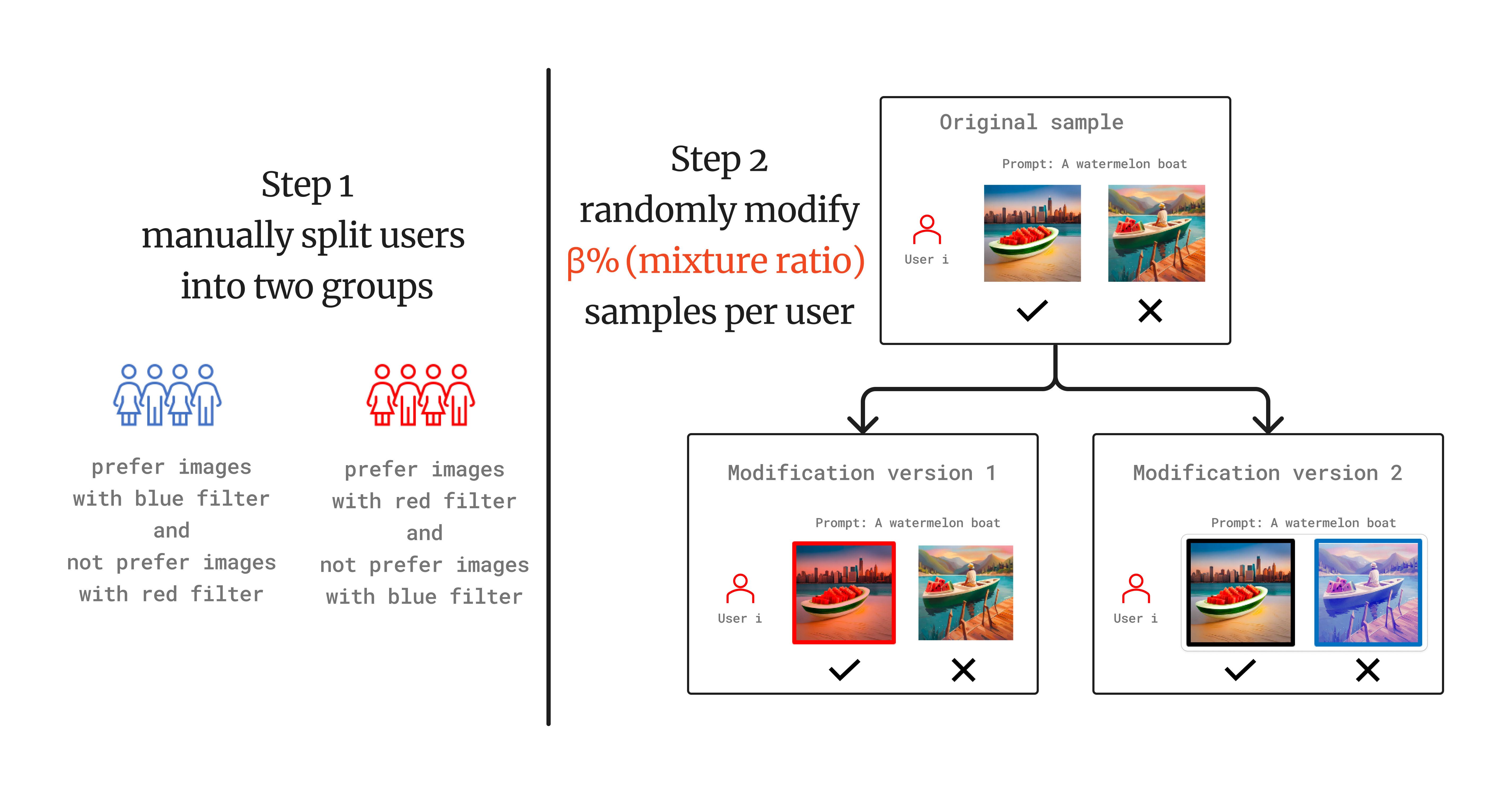}
    \caption{The construction diagram for the semi-synthetic Pick-a-Pic dataset. It involves randomly selecting approximately 100,000 samples from the Pick-a-Pic dataset and dividing the user IDs into two disjoint groups. We assume one group prefers images with ``cold'' (blue) filters and the other with ``warm'' (red) filters. To incorporate diverse color filter preferences, we randomly select $\beta\%$ of samples per user on which to apply filters.}
    \label{fig:semi-synthetic-dataset-construction}
\end{figure}

\begin{table}[]
\centering
\caption{Number of samples in each split of the newly constructed Pick-a-Filter dataset.}
\label{tab:pick-a-filter}
\vspace{3mm}
\begin{tabular}{@{}cl|ccc@{}}
\toprule
\multicolumn{2}{c}{Category}      & Train      & Val      & Test \\ \midrule
\multirow{3}{*}{Group 1} & Seen   & 58831      & 628      & 1597 \\
                         & Unseen & $~$9527 & $~$79 & 1886 \\
                         & Total  & 68358      & 707      & 3483 \\
\midrule
\multirow{3}{*}{Group 2} & Seen   & 57200      & 404      & 2096 \\
                         & Unseen & $~$9402 & $~$52 & 1812 \\
                         & Total  & 66602      & 456      & 3908 \\ 
\bottomrule 
\end{tabular}
\end{table}

\section{Experiment Details}

\subsection{Heterogeneous Synthetic Dataset}
\label{app:simu}

\textbf{Experiment Setup.} We introduce the dataset simulation procedure in the section \ref{subsec:numerical_simulation}. We use the following hyper-parameters to generate the synthetic dataset $d=16, K=3, N=100, n=100, \delta=1$. We generate another 50 comparison pairs per user as the held-out dataset. (Notice, we didn't simulate the prompt-guided item generation $\{x_c, x_l, x_r\}$ procedure. Instead, we directly draw the item $\{x_l,x_r\}$ from a normal distribution for simplicity.) In the experimental setup, we apply a toy version of the modeling design A, the distance between the synthetic item and the user ideal point is measured by $\|f(x)-f(u)\|_2$. We use a projection matrix (i.e. one-layer MLP network without bias term and activation function) as the model architecture. We randomly initialize the learnable parameters of prototypical user groups and user weights. We use Adam as the optimizer. The learning rate of the projector $f$ is $5e-4$. The learning rate of the learnable parameters of prototypical user groups and user weights is $5e-3$. The weight decay of the projection matrix $f$ is $1e-3$. To guarantee convergence, we run a total of 1000 epochs for each run. We run multiple trials to explore the influence of each factor: 1) varying the number of samples of seen users $n=\{20,40,60,80,100,400,800,1000\}$, $d=\{2,16\}$, $K=5$, $N=250$, 2) varying the number of samples of new users $n_{new}=\{5,10,20,30,40,50,100\}$, $d=\{2,16\}$,$K=5$, $n=50$, 3) varying the number of groups $K=\{2,3,4,5,6\}$, $d=\{2,16\}$, $n=50$, $N=50*K$.

\begin{figure}[ht]
    \centering
    \begin{minipage}[b]{0.32\textwidth}
        \centering
        \includegraphics[width=\linewidth]{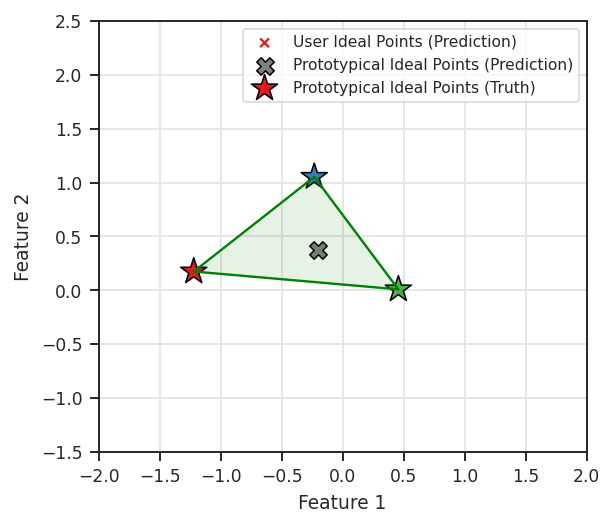}
            \caption{Partition setting\\ \# prototypes in model = 1}
        \label{fig:image1}
    \end{minipage}
    \hfill
    \begin{minipage}[b]{0.32\textwidth}
        \centering
        \includegraphics[width=\linewidth]{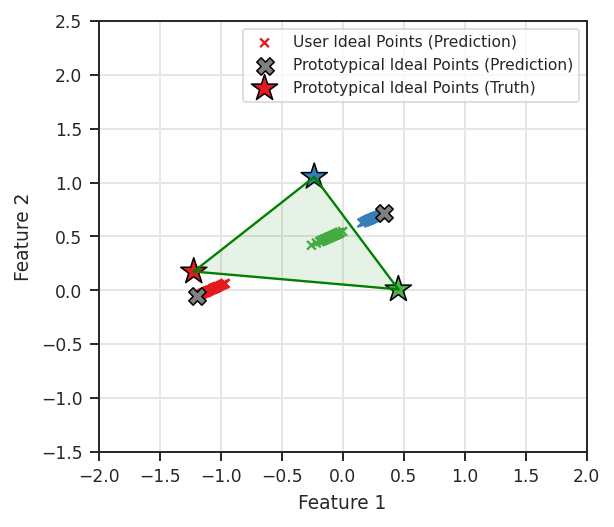}
        \caption{Partition setting\\ \# prototypes in model = 2}
        \label{fig:image2}
    \end{minipage}
    \hfill
    \begin{minipage}[b]{0.32\textwidth}
        \centering
        \includegraphics[width=\linewidth]{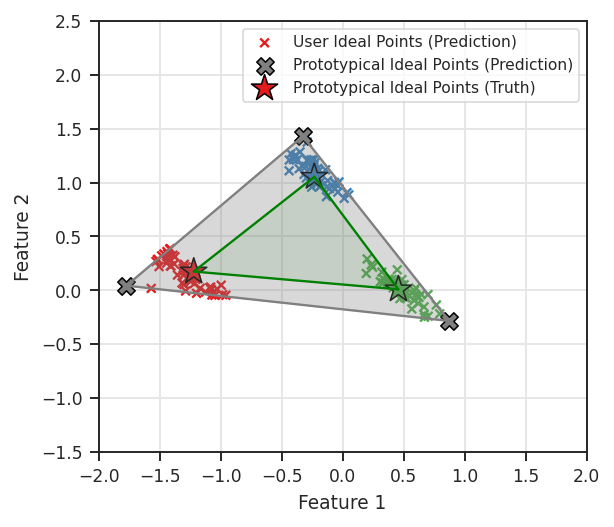}
        \caption{Partition setting\\ \# prototypes in model = 3}
        \label{fig:image3}
    \end{minipage}

    \begin{minipage}[b]{0.32\textwidth}
        \centering
        \includegraphics[width=\linewidth]{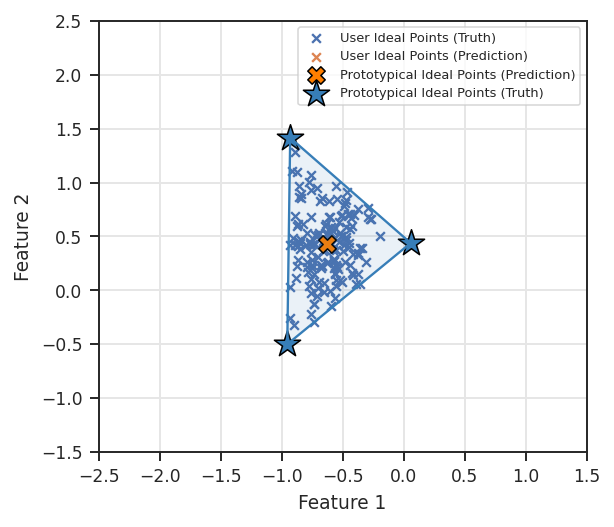}
        \caption{Mixture setting\\ \# prototypes in model = 1}
        \label{fig:image4}
    \end{minipage}
    \hfill
    \begin{minipage}[b]{0.32\textwidth}
        \centering
        \includegraphics[width=\linewidth]{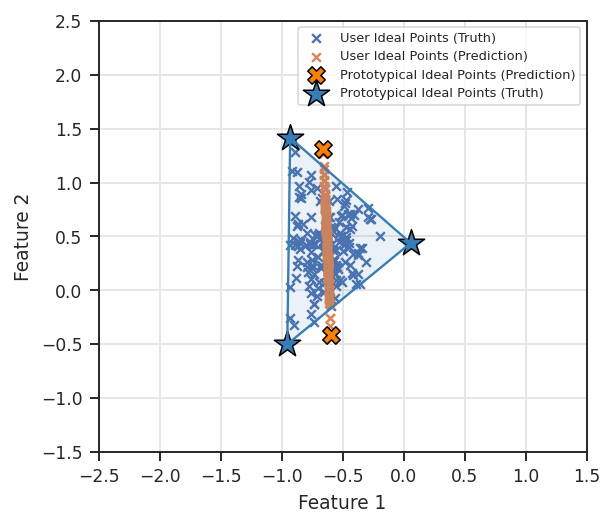}
        \caption{Mixture setting\\ \# prototypes in model = 2}
        \label{fig:image5}
    \end{minipage}
    \hfill
    \begin{minipage}[b]{0.32\textwidth}
        \centering
        \includegraphics[width=\linewidth]{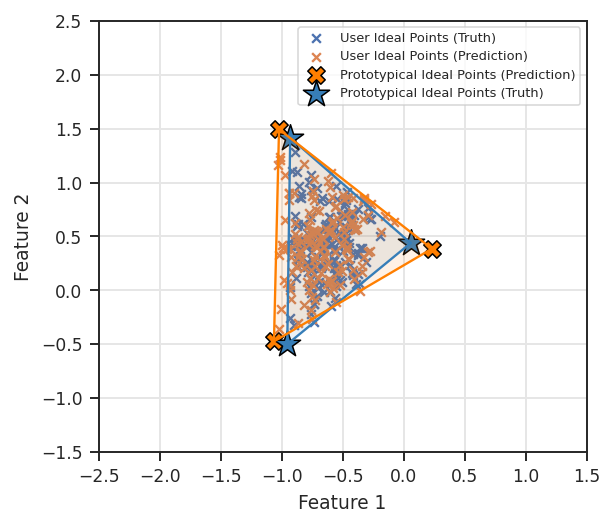}
        \caption{Mixture setting\\ \# prototypes in model = 3}
        \label{fig:image6}
    \end{minipage}
\end{figure}

\begin{figure}[ht]
    \centering
    \begin{subfigure}[b]{0.32\textwidth}
        \centering
        \includegraphics[width=\textwidth]{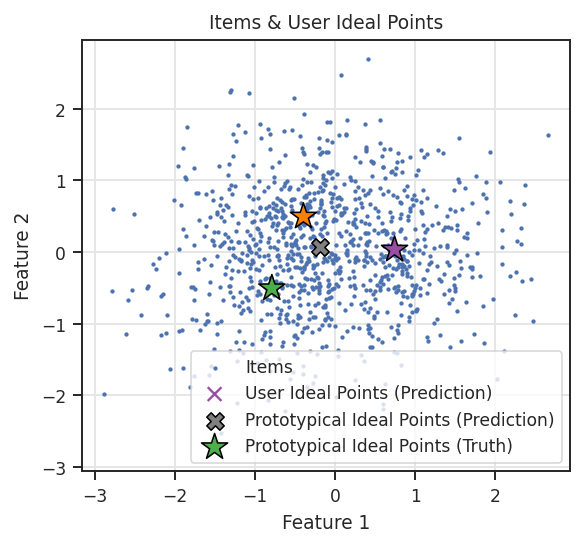}
        \caption{Test accuracy: 72.2\%}
        \label{fig:gaussian_synthetic_k1}
    \end{subfigure}
    \hfill 
    \begin{subfigure}[b]{0.32\textwidth}
        \centering
        \includegraphics[width=\textwidth]{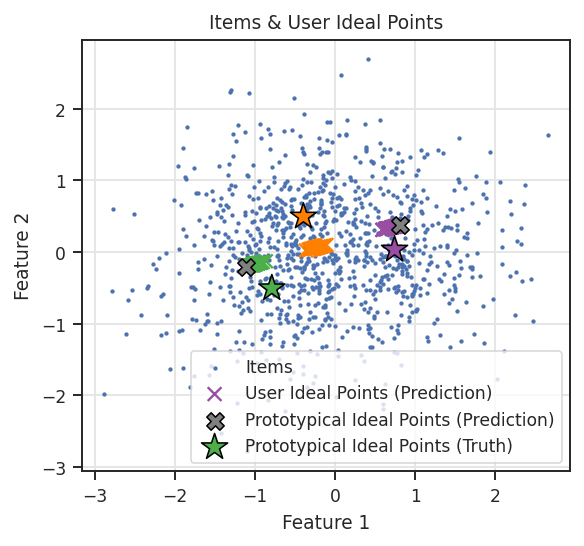}
        \caption{Test accuracy: 83.96\%}
        \label{fig:gaussian_synthetic_k2}
    \end{subfigure}
    \hfill 
    \begin{subfigure}[b]{0.32\textwidth}
        \centering
        \includegraphics[width=\textwidth]{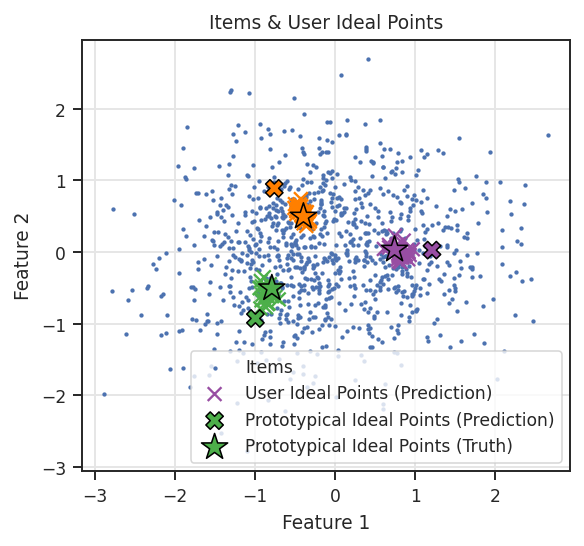}
        \caption{Test accuracy: 91.26\%}
        \label{fig:gaussian_synthetic_k3}
    \end{subfigure}
    \caption{Normally distributed items with $d = 2, K = 3, N = 100, n = 100$. This figure plots all items, the predicted user ideal points, and the true user ideal points in the feature space. Recall that in our modeling design, the distance between the user ideal point and the item reflects the user's preference; hence, the closer the predicted user ideal point is to the true ideal points, the higher the performance. As shown in the figures above, when we choose the hyperparameter $K=3$ (the correct number of groups), our model can accurately capture the group structure and predict each user's ideal points.}
    \label{fig:gaussian_synthetic_dataset}
\end{figure}

\subsection{Pick-a-Filter Dataset} \label{app:hetero-pick}
\textbf{Experiment Setup.} We choose two-layer MLP networks with ReLU activation and residual connection as the prompt mapping function $g_k$ and the output mapping function $f$. To avoid the overfitting issue, we set the dropout rate as $0.5$ and weight decay as $1e-2$. We use Adam optimizer with a $1e-4$ learning rate. When we measure the model's performance, we load the best checkpoint evaluated on the validation set.

\begin{figure}[t]
\includegraphics[width=0.6\linewidth]{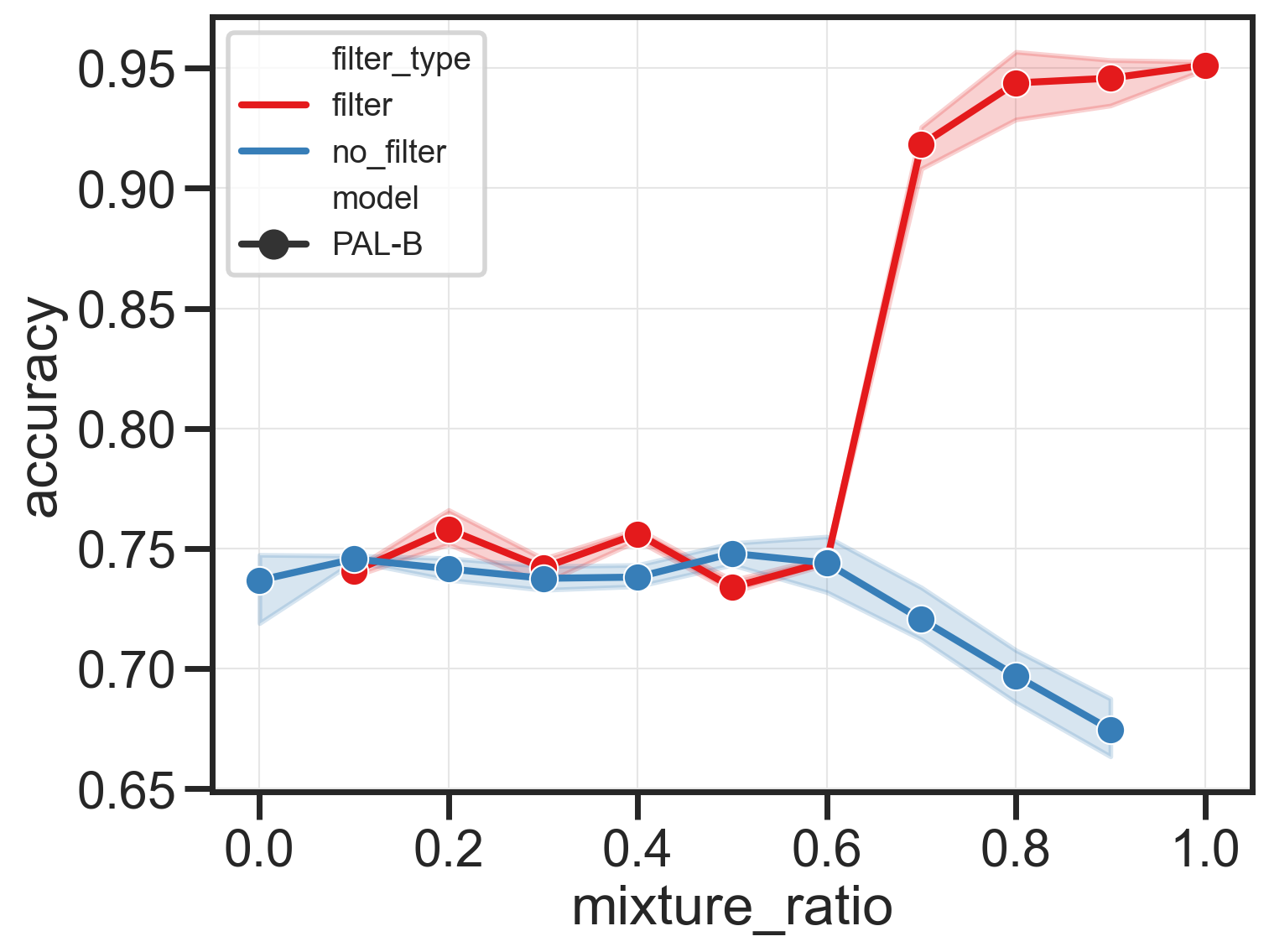}
\centering
\caption{Test accuracy on color-filtered or original pairs in \textit{Pick-a-Filter} dataset}
\label{fig:hetero-pickapic-mixture-ratio-seperate}
\end{figure}

\textbf{Results.} To check whether our model trained on \textit{Pick-a-Filter} dataset is capturing the users' preference features or is just remembering colors, we verify the test accuracy separately on the color-filtered pairs and original pairs in the mixture-ratio dataset. Figure~\ref{fig:hetero-pickapic-mixture-ratio-seperate} shows that compared to the CLIP-H14 $\sim65\%$ test accuracy, our model's performance on the original no-filter pairs is still above the baseline, which verifies that our model utilizes both the users' original preference and the "injected" heterogeneous color preference.

\subsection{Summary Dataset} \label{app:summary}

\paragraph{Dataset.} Reddit TL;DR summary dataset curated by \cite{stiennon2020learning} contains a series of preferences over summaries generated by language models. High-quality workers are hired by the authors to annotate their preferences over the summaries. Workers hired followed a rubric provided by the authors, who periodically fired those workers who did not meet their performance criteria.

For each pair of summaries $\bx_{\mathrm{left}}$ and $\bx_{\mathrm{right}}$, a worker $u$ determines if $\bx_{\mathrm{left}}$ is preferred or not. Moreover, each pair is also accompanied by the unique identifier of the worker who provides the preference. This would allow us to apply our model to such a dataset.

\paragraph{Experiment Setup and Results: Comparing to \cite{stiennon2020learning}} We trained our model A on the modified summary dataset with $K=1, \dots, 10$. This is because we want to evaluate the performance of generalization on the unseen users. We split the given testing set into a seen testing set and an unseen dataset, where the seen testing set contains users in the training set, and the unseen dataset contains only users that are not in the training set. The seen testing set is used to validate the performance of seen user, unseen comparison generalization. We are going to conduct a train, test split on the unseen dataset to evaluate the performance of unseen user, unseen comparison generalization.

We adopt the hyperparameters used in the experiment described in \ref{app:hetero-pick} in order to save time on hyperparameter tuning.

Table~\ref{table:summary} compares the performance of \PAL to the 1.7B reward model in~\cite{stiennon2020learning}. The overall accuracy is the weighted average of seen and unseen user accuracy. We want to emphasize that the main advantage of our model is that we do not require the existence of a supervised fine-tuned model. We used \texttt{all-mpnet-base-v2} sentence transformer~\cite{reimers-2019-sentence-bert}, which contains around 105M parameters, to generate the embedding for summaries and trained a 2-layer MLP with roughly 592K parameters.

\begin{table}[]
\centering
\caption{The performance of our method vs. the 1.3B reward model from~\cite{stiennon2020learning} on Summary Dataset.  Notably, our approach does not necessitate a supervised fine-tuned model. We leverage the \texttt{all-mpnet-base-v2} sentence transformer~\cite{reimers-2019-sentence-bert}, with 105M parameters, for summary embeddings, and train a 2-layer MLP, with 592K parameters.}
\vspace{2mm}
\begin{tabular}{@{}lcccc@{}}
\toprule
                     & $K=1$            & $K=2$            & $K=3$                   & $K=4$                \\ \midrule
Seen user accuracy   & $60.85 \pm 0.11$ & $60.95 \pm 0.12$ & $60.77 \pm 0.10$        & $60.81 \pm 0.12$     \\
Unseen user accuracy & $64.13 \pm 0.14$ & $64.18 \pm 0.19$ & $64.04 \pm 0.23$        & $63.99 \pm 0.12$     \\
Overall              & $61.36 \pm 0.12$ & $61.45 \pm 0.13$  & $61.28 \pm 0.13$        & $61.30 \pm 0.12$     \\ \midrule
                     & $K=5$            & $K=6$            & $K=7$                   & $K=8$                \\ \midrule
Seen user accuracy   & $60.91 \pm 0.10$ & $60.81 \pm 0.06$ & $60.71\pm 0.06$         & $60.88 \pm 0.13$     \\
Unseen user accuracy & $64.33 \pm 0.10$ & $64.11 \pm 0.15$ & $64.12 \pm 0.17$        & $64.12 \pm 0.13$     \\
Overall              & $61.44 \pm 0.10$ & $61.32 \pm 0.08$  & $61.25 \pm 0.09$        & $61.38 \pm 0.13$     \\ \midrule
                     & $K=9$            & $K=10$           & Stiennon et. al. (1.3B) & \multicolumn{1}{l}{} \\ \midrule
Seen user accuracy   & $60.95 \pm 0.10$ & $60.93 \pm 0.12$ & -                       & \multicolumn{1}{l}{} \\
Unseen user accuracy & $64.07 \pm 0.20$ & $64.19 \pm 0.11$ & -                       & \multicolumn{1}{l}{} \\
Overall              & $61.43 \pm 0.12$ & $61.44 \pm 0.12$ & $65.80 \pm 2.00$        &                      \\ \bottomrule
\end{tabular}
\label{table:summary}
\end{table}

\paragraph{Experiment Setup and Results: Comparing to \cite{li2024personalized}}

We evaluate our model A with hinge loss on a trimmed version of the summary dataset described in \cite{li2024personalized}, to compare our results with theirs. In \cite{li2024personalized}, the original training set of the summary dataset is filtered with summaries generated by SFT policies and only those comparisons made by the top 10 workers who conduct the most pairwise comparisons are kept. The test dataset is split into 2 folds where those comparisons made by the 10 workers are used to evaluate the generalization performance on seen users, whereas those comparisons made by other workers are used to evaluate the generalization performance on unseen users. 

Table \ref{table:summary_main} compares the performance of the method to the one proposed in \cite{li2024personalized}. We use the weighted average of prototypes learned as the general ideal point for new users to conduct zero-shot learning. We emphasize that even though our model only has 594K parameters and the sentence embeddings we used are generated from \texttt{all-mpnet-base-v2} sentence transformer \cite{reimers-2019-sentence-bert}, which contains around 105M parameters, we still can achieve on par performance, especially in terms of unseen accuracy.



\section{Modeling Design} \label{app:model}

\begin{algorithm}
\caption{\PAL-A algorithm}
\begin{algorithmic}[1]
\Require Dataset $\mathcal{D} = \left\{ \{ (\bxl, \bxr; \bxc)^{(i)}_j\}_{j=1}^{m_i} \right\}_{i=1}^N$, loss function $\ell$, model class for $f_{\theta}$, prototypes $\bP=[\bp_1, ..., \bp_K]$, $\bp_k \in \mathbb{R}^d$, user weights $\bW = [\bw^{(1)}, ..., \bw^{(N)}]$, where $\bw^{(i)} \in \Delta^{K}$.
\For{each iteration}
    \State \textbf{sample} a mini-batch $\left\{ (\bxl, \bxr; \bxc)^{(i)}_j \right\}$ \Comment{random pairs, not ordered by users}
       \State \textbf{User Ideal Points:}  $\ba^{(i)} = \bP \cdot \bw^{(i)}$
        \State \textbf{Distances}:
        \State \quad $d_{l,j}^{(i)}=||f_\theta\left(\bx_{l,j}^{(i)};\bx_{c,j}^{(i)}\right) - f_\theta(\ba^{(i)})||^2_2$, \quad $d_{r,j}^{(i)}=||f_\theta\left(\bx_{r,j}^{(i)};\bx_{c,j}^{(i)}\right)-f_\theta(\ba^{(i)})||^2_2$
        \State \textbf{Loss}:  $\ell_j^{(i)}(\bx_{l,j}^{(i)}, \bx_{r,j}^{(i)} ; \bx_{c,j}^{(i)}) = \ell(d_{r,j}^{(i)}-d_{l,j}^{(i)})$
    \State \textbf{Update Step:} $\mathbf{arg max}_{\theta,\bP,\{\bw^{(i)} \in \Delta^{K}\}_{i=1}^N} \sum_{i, j} l^{(i)}_j(\bx_{l,j}^{(i)}, \bx_{r,j}^{(i)}; \bx_{c,j}^{(i)}) $ 
\EndFor
\end{algorithmic}
\end{algorithm}

\begin{algorithm}
\caption{\PAL-B algorithm}
\begin{algorithmic}[1]
\Require Preference data $\mathcal{D} = \left\{ \{ (\bxl, \bxr; \bxc)^{(i)}_j\}_{j=1}^{m_i} \right\}_{i=1}^N$, loss function $\ell$, mapping function $f_{\theta}$, prototype mapping functions $\{g_{\theta_k}\}_{k=1}^K$, user weights $\{\bw^{(i)}:= [w^{(i)}_1, ..., w^{(i)}_K]\}_{i=1}^N$.

\For{each iteration}
    \State \textbf{sample} a mini-batch $\left\{ (\bxl, \bxr; \bxc)^{(i)}_j \right\}$ \Comment{random pairs, not ordered by users}
       \State \textbf{User Ideal Point \small{(condition on prompts)}}:
       \State \quad $\ba^{(i)} = \left[g_{\theta_i}(\bx_{c,j}^{(i)}),...,g_{\theta_K}(\bx_{c,j}^{(i)})\right]^\top \cdot \bw^{(i)}$
        \State \textbf{Distance}:
            \State \quad $d_{l,j}^{(i)}=\langle f_\theta\left(\bx_{l,j}^{(i)}\right), \ba^{(i)}\rangle$, \quad $d_{r,j}^{(i)}=\langle f_\theta\left(\bx_{r,j}^{(i)}\right), \ba^{(i)}\rangle$
        \State \textbf{Loss}: $\ell_j^{(i)}(\bx_{l,j}^{(i)}, \bx_{r,j}^{(i)} ; \bx_{c,j}^{(i)}) = \ell(d_{r,j}^{(i)}-d_{l,j}^{(i)} )$
    \State \textbf{Update Step:} $\mathbf{argmax}_{\Theta,\bP,\{\bw^{(i)}\in \Delta^{K}\}_{i=1}^N}\sum \ell_{j}^{(i)}(\bx_{l,j}^{(i)},\bx_{r,j}^{(i)} ; \bx_{c,j}^{(i)}) $ 
\EndFor
\end{algorithmic}
\end{algorithm}

\section{Broader Impacts}

This paper presents novel contributions to the field of machine learning towards foundations for learning from heterogeneous preferences aiding the development of models and algorithms to move the needle towards plurality. 

%% file: figures/models-diagram.tex
\begin{figure}[t]
    \centering
    \includegraphics[width=\textwidth]{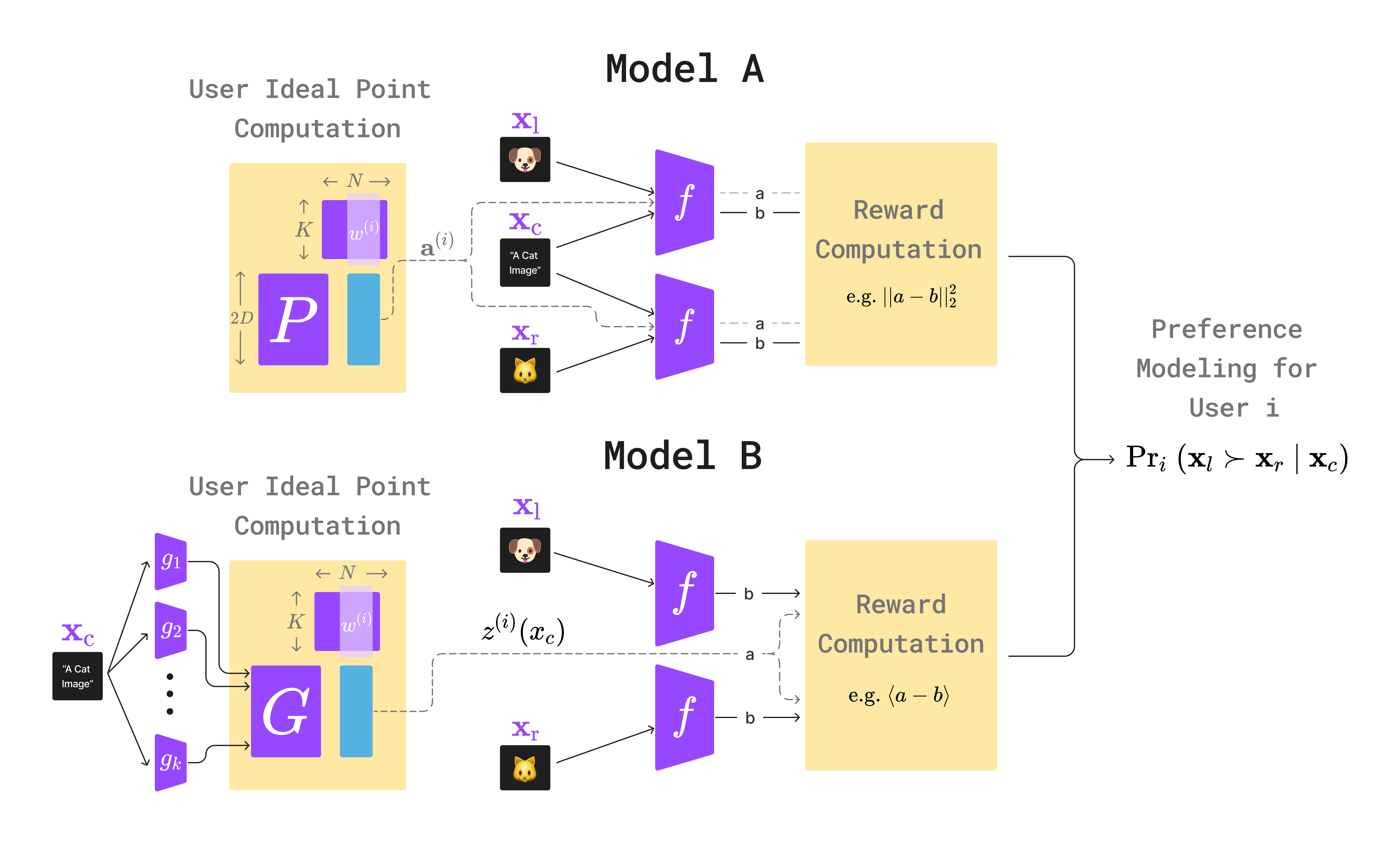}
    \caption{Illustration of \PAL framework for learning from diverse preferences (Section~\ref{sec:models}). For any user $i$, the probability of preferring $\bxl$ to $\bxr$ for the context $\bx_c$ is computed by a reward model $\rtheta^{(i)}$ which uses a mixture modeling approach to assign a scalar reward to a sample (e.g. $\bxl$ or $\bxr$) given context ($\bxc$). In \PAL-A, each user $i$'s preference $a^{(i)}$ is modeled as a convex combination of $K$ prototypical preferences, i.e. $a^{(i)} = Pw^{(i)}$. In \PAL-B, each user $i$'s preference $z^{(i)}(\bxc)$ is modeled as a convex combination of $K$ prototypical functions $g_1\cdot\cdot\cdot g_K$, i.e. $z^{(i)}(\bxc) = $. Reward function formulated using \PAL framework can be used flexibly, e.g., with fixed preference points (Model A), with preference points that are functions of the context/prompt $\bxc$ (Model B).
    }
    \label{fig:model_arch}
\end{figure}